\documentclass[twoside,11pt]{article}

\usepackage{jmlr2e}
\usepackage[dvips]{color}
\usepackage{amssymb}
\usepackage{amsmath}
\usepackage{graphicx}
\usepackage{algorithm}
\usepackage{algorithmic}
\newcommand{\st}{\mathop{\rm s.t.}}
\newcommand{\mb}[1]{\mathbf{#1}}

\newcommand{\oldversion}[1]{}
\newcommand{\newversion}[1]{#1}

\renewcommand{\oldversion}[1]{}
\renewcommand{\newversion}[1]{#1}

\ShortHeadings{Robustness and Regularization of SVMs}{Xu, Caramanis
and Mannor} \firstpageno{1}

\begin{document}

\title{Robustness and Regularization of Support Vector Machines}
\author{\name Huan Xu \email xuhuan@cim.mcgill.ca,
\\
\addr Department of Electrical and Computer Engineering, McGill
University, Canada \AND \name Constantine Caramanis \email
cmcaram@ece.utexas.edu\\ \addr Department of Electrical and Computer
Engineering, The University of Texas at Austin,  USA \AND
 \name Shie Mannor\email shie.mannor@mcgill.ca\\ \addr Department of Electrical and Computer Engineering, McGill
University, Canada}

\editor{Alexander Smola} \maketitle

\begin{abstract}
We consider regularized support vector machines (SVMs)
and show that they are precisely equivalent to a new robust
optimization formulation. We show that this equivalence of
robust optimization and regularization has implications for both
algorithms, and analysis. In terms of algorithms, the equivalence 
suggests more general SVM-like algorithms for classification that explicitly build
in protection to noise, and at the same time control overfitting. 
On the analysis front, the equivalence of robustness and regularization, 
provides a robust optimization interpretation for the success of regularized SVMs.
We use the this new robustness interpretation of SVMs to give a new proof
of consistency of (kernelized) SVMs, thus establishing robustness as the {\em reason}
regularized SVMs generalize well.

\end{abstract}

\section{Introduction}

Support Vector Machines (SVMs for short) originated in
\citet{Boser92} and can be traced back to as early as
\citet{Vapnik63} and \citet{Vapnik74}. They continue to be one of
the most successful algorithms for classification. SVMs address the
classification problem by finding the hyperplane in the feature
space that achieves maximum sample margin when the training samples
are separable, which leads to minimizing the norm of the classifier.
When the samples are not separable, a penalty term that approximates
the total training error is considered \citep{Bennett92,Cortes95}.
It is well known that minimizing the training error itself can lead
to poor classification performance for new unlabeled data; that is,
such an approach may have poor generalization error because of,
essentially, overfitting \citep{Vapnik91}. A variety of
modifications have been proposed to combat this problem, one of the
most popular methods being that of minimizing a combination of the
training-error and a regularization term. The latter is typically
chosen as a norm of the classifier. The resulting regularized
classifier  performs better on new data. This phenomenon is often
interpreted from a statistical learning theory view: the
regularization term restricts the complexity of the classifier,
hence the deviation of the testing error and the training error is
controlled \citep[see][and references
therein]{Smola98,Evgeniou00,Bartlett02,Koltchinskii02,Bartlett05}.

In this paper we consider a different setup,  assuming that the
training data are generated by the true underlying distribution, but
some non-i.i.d. (potentially adversarial) disturbance is then added to
the samples we observe. We follow a robust optimization
\citep[see][and references therein]{Ghaoui97, Ben-tal99,
Bertsimas04} approach, i.e., minimizing the worst possible empirical
error under such disturbances. The use of robust optimization in
classification is not new \citep[e.g.,][]{Shivaswamy06,
Bhattacharyya04b, Lanckriet02}. Robust classification models studied
in the past have considered only box-type uncertainty sets, which allow
the possibility that the data have all been skewed in some
non-neutral manner by a correlated disturbance. This has made it
difficult to obtain non-conservative generalization bounds.
Moreover, there has not been an explicit connection to the
regularized classifier, although at a high-level it is known that
regularization and robust optimization are related
\citep[e.g.,][]{Ghaoui97,Anthony99}. The main contribution in this
paper is solving the robust classification problem for a class of
non-box-typed uncertainty sets, and providing a linkage between
robust classification and the standard regularization scheme of
SVMs. In particular, our contributions include the following:
\begin{itemize}
\item We solve the robust SVM formulation for a class of non-box-type uncertainty sets.
This permits finer control of the
adversarial disturbance, restricting it to satisfy aggregate
constraints across data points, therefore reducing the possibility
of highly correlated disturbance.
\item We show that the standard regularized SVM classifier is a special
case of our robust classification, thus explicitly relating
robustness and regularization. This provides an alternative
explanation to the success of regularization, and also suggests new
physically motivated ways to construct regularization terms.
\item We relate our robust formulation to several probabilistic
formulations. We consider a chance-constrained  classifier (i.e., a classifier with probabilistic constraints
on misclassification) and show that our robust formulation can approximate
it far less conservatively than previous robust formulations could possibly do.
We also consider a Bayesian setup, and show that this can be used to provide
a principled means of selecting the regularization coefficient without cross-validation.
\item We show that the robustness perspective, stemming from a non-i.i.d. analysis, can
be useful in the standard learning (i.i.d.) setup, by using it  to  prove
consistency for standard SVM classification, {\it without using VC-dimension
or stability arguments}. This result
implies that generalization ability is a direct result of robustness
to local disturbances; it therefore suggests a new justification for good performance,
and consequently allows us to construct learning algorithms that
generalize well by robustifying non-consistent algorithms.
\end{itemize}

\subsubsection*{Robustness and Regularization}
We comment here on the explicit equivalence of robustness and
regularization. We briefly explain how this observation is different
from previous work and why it is interesting. Certain equivalence
relationships between robustness and regularization have been
established for problems other than classification
\citep{Ghaoui97,Ben-tal99,Bishop95}, but their results do not
directly apply to the classification problem. Indeed, research on
classifier regularization mainly discusses its effect on bounding
the complexity of the function class
\citep[e.g.,][]{Smola98,Evgeniou00,Bartlett02,Koltchinskii02,Bartlett05}.
Meanwhile, research on robust classification has not attempted to
relate robustness and regularization \citep[e.g.,][]{Lanckriet02,
Bhattacharyya04,
Bhattacharyya04b,Shivaswamy06,Trafalis07,Globerson06}, in part due
to the robustness formulations used in those papers.
In fact, they all consider robustified versions of {\em regularized}
classifications.\footnote{\citet{Lanckriet02} is perhaps the only
exception, where a regularization term is added to the covariance
estimation rather than to the objective function.}
\citet{Bhattacharyya04c} considers a robust formulation for
box-type uncertainty, and relates this robust formulation with
regularized SVM. However, this formulation involves a non-standard
loss function that does not bound the $0-1$ loss, and hence its
physical interpretation is not clear.

The connection of robustness and regularization in the SVM context
is important for the following reasons. First, it gives an
alternative and potentially powerful explanation of the
generalization ability of the regularization term. In the classical
machine learning literature, the regularization term bounds the
complexity of the class of classifiers. The robust view of
regularization regards the testing samples as a perturbed copy of
the training samples. We show that when the total perturbation
is given or bounded, the regularization term bounds the gap between
the classification errors of the SVM on these two sets of samples.
In contrast to the standard PAC approach, this bound depends neither
on how rich the class of candidate classifiers is, nor on an
assumption that all samples are picked in an i.i.d. manner. In
addition, this suggests novel approaches to designing good
classification algorithms, in particular, designing the
regularization term. In the PAC structural-risk minimization
approach, regularization is chosen to minimize a bound on the
generalization error based on the training error and a complexity
term. This complexity term typically leads to overly emphasizing the
regularizer, and indeed this approach is known to often be too
pessimistic~\citep{Kearns97} for problems with more structure.
The robust approach offers another
avenue. Since both noise and robustness are physical processes, a
close investigation of the application and noise characteristics at
hand, can provide insights into how to properly robustify, and
therefore regularize the classifier. For example, it is known that
normalizing the samples so that the variance among all features is
roughly the same (a process commonly used to eliminate the scaling
freedom of individual features) often leads to good generalization
performance. From the robustness perspective, this simply says that
the noise is anisotropic (ellipsoidal) rather than spherical, and
hence an appropriate robustification must be designed to fit this
anisotropy.

We also show that using the robust optimization viewpoint, we obtain
some probabilistic results outside the PAC setup. In
Section~\ref{sss.probinter} we bound the probability that a noisy
training sample is correctly labeled. Such a bound considers the
behavior of {\em corrupted} samples and is hence different from the
known PAC bounds. This is helpful when the training samples and the
testing samples are drawn from different distributions, or some
adversary manipulates the samples to prevent them from being
correctly labeled (e.g., spam senders change their patterns from
time to time to avoid being labeled and filtered).
Finally, this connection of
robustification and regularization also provides us with new proof
techniques as well (see Section~\ref{sss.consistency}).

We need to point out that there are several different definitions of
robustness in literature. In this paper, as well as the
aforementioned robust classification papers, robustness is mainly
understood from a Robust Optimization perspective, where a min-max
optimization is performed over all possible disturbances. An
alternative interpretation of robustness stems from the rich
literature on Robust Statistics
\citep[e.g.,][]{Huber81,HampelRonchettiRousseeumStahel86,RousseeuwLeroy87,MaronnaMartinYohai06},
which studies how an estimator or algorithm behaves under a small
perturbation of the statistics model. For example, the
Influence Function approach, proposed in \citet{Hample74} and
\citet{HampelRonchettiRousseeumStahel86}, measures the impact  of an
infinitesimal amount of contamination of the original
distribution on the quantity of interest. Based on this notion of
robustness, \citet{ChristmannSteinwart04} showed that many kernel
classification algorithms, including SVM, are robust in the sense of
having a finite Influence Function. A similar result for regression
algorithms is shown in \citet{ChristmannSteinwart07} for smooth loss
functions, and in \citet{ChristmannMessem08} for non-smooth loss
functions where a relaxed version of the Influence Function is applied.
In the machine learning literature, another widely used notion closely
related to robustness is the {\em stability}, where an algorithm is
required to be robust (in the sense that the output function does
not change significantly) under a specific perturbation: deleting
one sample from the training set. It is now well known that a stable
algorithm such as SVM has desirable generalization properties, and
is statistically consistent under mild technical conditions; see for
example
\citet{Bousquet02,KutinNiyogi02,PoggioRifkinMukherjeeNiyogi04,MukherjeeNiyogiPoggioRifkin06}
for details. One main difference between Robust Optimization and
other robustness notions is that the former is constructive rather
than analytical. That is, in contrast to robust statistics or the
stability approach that measures the robustness of a {\em given}
algorithm, Robust Optimization can {\em robustify} an algorithm: it
 converts a given algorithm to a robust one. For
example, as we show in this paper, the RO version of a naive
empirical-error minimization is the well known SVM. As a
constructive process, the RO approach also leads to additional
flexibility in algorithm design, especially when the nature of the
perturbation is known or can be well estimated.

{\bf Structure of the Paper:} This paper is organized as follows. In
Section~\ref{sec.robust} we  investigate the correlated disturbance
case, and  show the equivalence between the robust classification
and the regularization process. We develop the connections to
probabilistic formulations in Section~\ref{sss.probinter}, and
prove a consistency result based on robustness analysis in
Section~\ref{sss.consistency}. The kernelized version is
investigated in Section~\ref{sec.kernelization}.  Some concluding
remarks are given in Section~\ref{sec.conclude}.

{\bf Notation:} Capital letters are used to denote matrices, and
boldface letters are used to denote column vectors. For a given norm
$\|\cdot\|$, we use $\|\cdot\|^*$ to denote its dual norm, i.e.,
$\|\mathbf{z}\|^*\triangleq \sup\{\mathbf{z}^\top
\mathbf{x}|\|\mathbf{x}\|\leq 1\}$. For a vector $\mathbf{x}$ and a
positive semi-definite matrix $C$ of the same dimension,
$\|\mathbf{x}\|_C$ denotes $\sqrt{\mathbf{x}^\top C\mathbf{x}}$. We
use $\mb{\delta}$ to denote disturbance affecting the samples. We
use superscript $r$ to denote the true value for an uncertain
variable, so that $\boldsymbol{\delta}_i^r$ is the true (but
unknown) noise of the $i^{th}$ sample. The set of non-negative
scalars is denoted by $\mathbb{R}^+$. The set of integers from $1$
to $n$ is denoted by $[1:n]$.

\section{Robust Classification and Regularization}\label{sec.robust}
\newversion{We consider the standard  binary classification
problem, where we are given a finite number of training samples
$\{\mathbf{x}_i, y_i\}_{i=1}^m \subseteq \mathbb{R}^n \times \{-1,
+1\}$, and must find a linear classifier, specified by the function
$h^{\mathbf{w},b}(\mathbf{x})=\mathrm{sgn}(\langle\mathbf{w},\,\mathbf{x}\rangle+b).$
For the standard regularized classifier, the parameters
$(\mathbf{w}, b)$ are obtained by solving the following convex
optimization problem:
\begin{eqnarray*}
\min_{\mathbf{w},b, \boldsymbol{\xi}}: && r(\mathbf{w},b) +\sum_{i=1}^m \xi_i \\
\st: &&  \xi_i \geq \big[1-y_i(\langle\mathbf{w}, \mathbf{x}_i\rangle+b)] \\
&& \xi_i \geq 0,
\end{eqnarray*}
where $r(\mathbf{w},b)$ is a regularization term. This is equivalent
to
\begin{equation*}
\min_{\mathbf{w},b}\left\{ r(\mathbf{w},b) +\sum_{i=1}^m
\max\big[1-y_i(\langle\mathbf{w}, \mathbf{x}_i\rangle+b),
0\big]\right\}.
\end{equation*} Previous robust classification work
\citep{Shivaswamy06,Bhattacharyya04,Bhattacharyya04b,Bhattacharyya04c,Trafalis07}
considers the classification problem where the input are subject to
(unknown) disturbances $\boldsymbol{\vec{\delta}} =
(\boldsymbol{\delta}_1,\dots,\boldsymbol{\delta}_m)$ and
 essentially solves the following min-max problem:
\begin{equation}
\label{eq:boxrobust}
\min_{\mathbf{w},b}\max_{\boldsymbol{\vec{\delta}}\in
\mathcal{N}_{\mbox{\tiny{box}}}} \left\{r(\mathbf{w},b)
+\sum_{i=1}^m \max\big[1-y_i(\langle\mathbf{w},\,
\mathbf{x}_i-\boldsymbol{\delta}_i\rangle+b), 0\big]\right\},
\end{equation} for a box-type uncertainty set $\mathcal{N}_{\mbox{\tiny{box}}}$. That is, let ${\cal N}_i$ denotes the
projection of $\mathcal{N}_{\mbox{\tiny{box}}}$ onto the
$\boldsymbol{\delta}_i$ component, then
$\mathcal{N}_{\mbox{\tiny{box}}} = {\cal N}_1 \times \cdots \times
{\cal N}_m$. Effectively, this allows simultaneous worst-case
disturbances across many samples, and  leads to overly conservative
solutions. The goal of this paper is to obtain a robust formulation
where the disturbances $\{\boldsymbol{\delta}_i\}$ may be
meaningfully taken to be correlated, i.e., to solve for a
non-box-type $\mathcal{N}$:
\begin{equation}\label{equ.robustclassifierformulation}
\min_{\mathbf{w},b}\max_{\boldsymbol{\vec{\delta}}\in \mathcal{N}}
\left\{r(\mathbf{w},b) +\sum_{i=1}^m
\max\big[1-y_i(\langle\mathbf{w},
\mathbf{x}_i-\boldsymbol{\delta}_i\rangle+b), 0\big]\right\}.
\end{equation} }
We briefly explain here the four reasons that motivate this ``robust
to perturbation'' setup and in particular the min-max form
of~(\ref{eq:boxrobust}) and~(\ref{equ.robustclassifierformulation}).
First, it can explicitly incorporate prior problem knowledge of
local invariance \citep[e.g,][]{TeoGlobersonRoweisSmola07}. For
example, in vision tasks, a desirable classifier should provide a
consistent answer if an input image slightly changes. Second,
there are situations where some adversarial opponents (e.g., spam
senders) will manipulate the testing samples  to avoid being
correctly classified, and the robustness toward such manipulation
should be taken into consideration in the training process
\citep[e.g,][]{Globerson06}. Or alternatively, the training samples
and the testing samples can be obtained from different processes and
hence the standard i.i.d. assumption is violated
\citep[e.g,][]{Bi04}. For example in real-time applications, the
newly generated samples are often less accurate due to time
constraints. Finally, formulations based on chance-constraints
\citep[e.g.,][]{Bhattacharyya04b,Shivaswamy06} are mathematically
equivalent to such a min-max formulation.

\begin{figure}[h]
\begin{center}
\begin{tabular}{cccc}
  \includegraphics[height=3cm, width=0.2\linewidth]{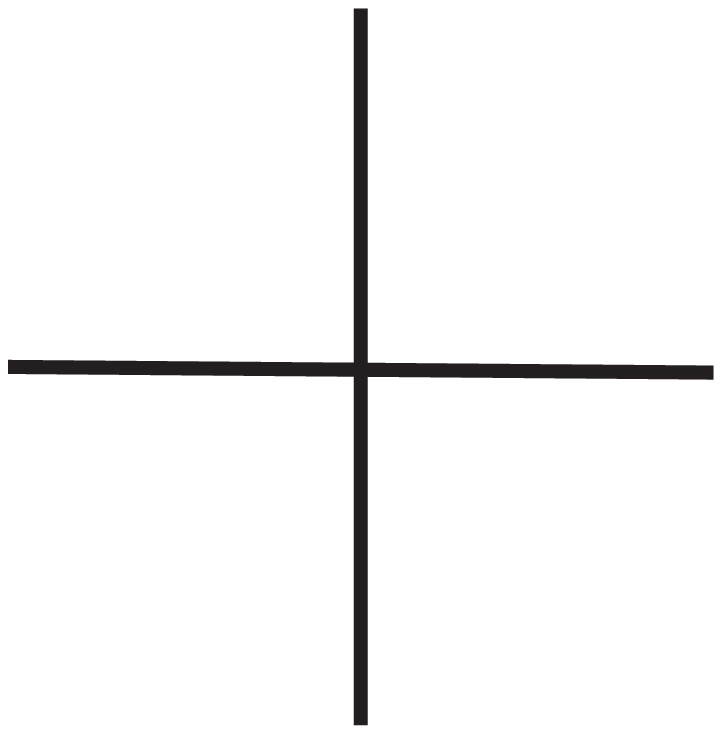} &
\includegraphics[height=3cm, width=0.2\linewidth]{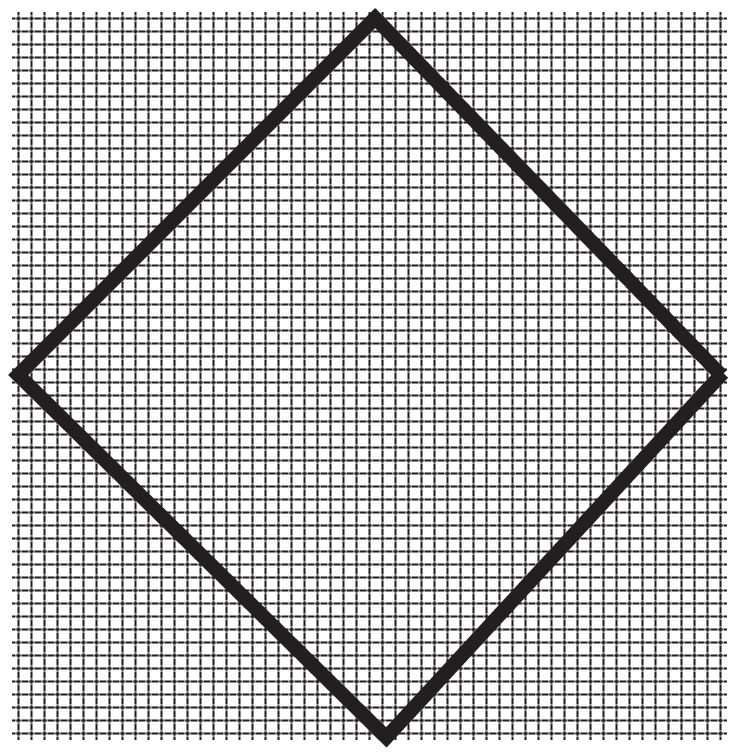} &
\includegraphics[height=3cm, width=0.2\linewidth]{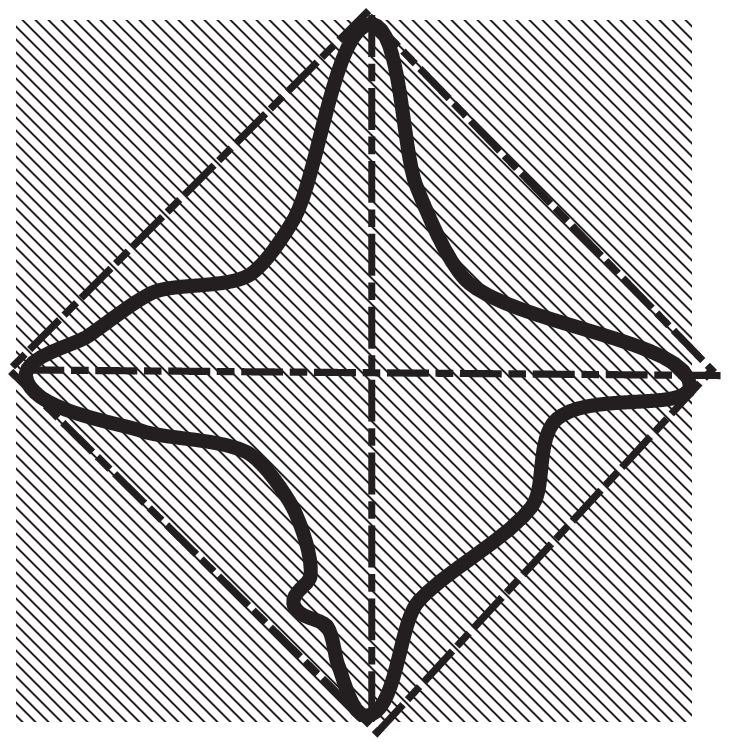} &
\includegraphics[height=3cm, width=0.2\linewidth]{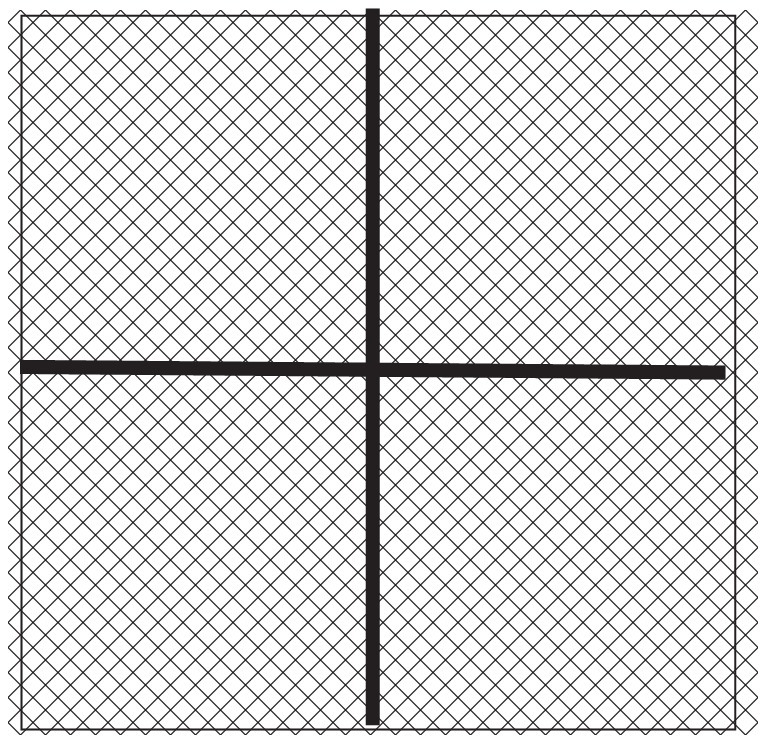}\\
a. $\mathcal{N}^-$ & b. $\mathcal{N}^+$ & c. $\mathcal{N}$ & d. Box
uncertainty
\end{tabular}\caption{Illustration of a Sublinear Aggregated Uncertainty Set
$\mathcal{N}$.}\label{fig.ccus}
\end{center}
\end{figure}

 We define
explicitly the correlated disturbance (or uncertainty) which we
study below.
\begin{definition}\label{def.concavecorrelate1}
A set $\mathcal{N}_0\subseteq \mathbb{R}^n$ is called an {\em Atomic
Uncertainty Set} if
\begin{equation*}
\begin{split}\mbox{(I)}\quad&\mathbf{0}\in
\mathcal{N}_0;\\ \mbox{(II)}\quad& \mbox{ For any }\mathbf{w}_0\in
\mathbb{R}^n: \sup_{\boldsymbol{\delta}}
[\mathbf{w}_0^\top\boldsymbol{\delta}]=\sup_{\boldsymbol{\delta}'}
[-\mathbf{w}_0^\top\boldsymbol{\delta}']<+\infty.
\end{split}\end{equation*}
\end{definition}
We use ``$\sup$'' here because the maximal value is not necessary
attained since $\mathcal{N}_0$ may not be a closed set. The second
condition of Atomic Uncertainty set basically says that the
uncertainty set is bounded and symmetric. In particular, all norm
balls and ellipsoids centered at the origin are atomic uncertainty
sets, while an arbitrary polytope might not be an atomic uncertainty
set.
\begin{definition}\label{def.concavecorrelate2}
 Let $\mathcal{N}_0$ be an atomic uncertainty set. A set
$\mathcal{N} \subseteq \mathbb{R}^{n\times m}$ is called a {\em
Sublinear Aggregated Uncertainty Set} of $\mathcal{N}_0$, if
\[\mathcal{N}^-\subseteq \mathcal{N}\subseteq \mathcal{N}^+,\]
\begin{equation*}
\begin{split}
\mbox{where:}\quad\mathcal{N}^- &\triangleq \bigcup_{t=1}^m
\mathcal{N}_t^-;\quad\quad\mathcal{N}_t^- \triangleq
\{(\boldsymbol{\delta}_1, \cdots,
\boldsymbol{\delta}_m)|\boldsymbol{\delta}_t \in \mathcal{N}_0;\, \,
\boldsymbol{\delta}_{i\not= t}=\mathbf{0}\}.\\
\mathcal{N}^+ &\triangleq  \{(\alpha_1\boldsymbol{\delta}_1,\cdots,
\alpha_m\boldsymbol{\delta}_m)| \sum_{i=1}^m
\alpha_i=1;\,\,\alpha_i\geq 0,\,\, \boldsymbol{\delta}_i\in
\mathcal{N}_0,\, i=1,\cdots, m\}.
\end{split}
\end{equation*}
\end{definition}
The Sublinear Aggregated Uncertainty definition models the case
where the disturbances on each sample are treated identically, but
their aggregate behavior across multiple samples is controlled. Some
interesting examples include
\begin{eqnarray*}
&&(1)\quad
\{(\boldsymbol{\delta}_1,\cdots,\boldsymbol{\delta}_m)|\sum_{i=1}^m
\|\boldsymbol{\delta}_i\| \leq c\}; \\
&&(2)\quad
\{(\boldsymbol{\delta}_1,\cdots,\boldsymbol{\delta}_m)|\exists t\in
[1:m];\, \|\boldsymbol{\delta}_t\| \leq c;\,\,
\boldsymbol{\delta}_{i}=\mathbf{0},\forall i\not=t\};\\
&&(3)\quad
\{(\boldsymbol{\delta}_1,\cdots,\boldsymbol{\delta}_m)|\sum_{i=1}^m
\sqrt{c\|\boldsymbol{\delta}_i\|} \leq c\}.
\end{eqnarray*}
All these examples have the same atomic uncertainty set ${\cal N}_0
= \big\{\boldsymbol{\delta}\big|\,\|\boldsymbol{\delta}\|\leq
c\big\}$. Figure~\ref{fig.ccus} provides an illustration of a
sublinear aggregated uncertainty set for $n=1$ and $m=2$, i.e., the
training set consists of two univariate samples.
\begin{theorem}\label{thm.robustequiv}
Assume $\{\mathbf{x}_i, y_i\}_{i=1}^m$ are non-separable,
$r(\cdot):\mathbb{R}^{n+1}\rightarrow \mathbb{R}$ is an arbitrary
function,  $\mathcal{N}$ is a Sublinear Aggregated Uncertainty set
with corresponding atomic uncertainty set $\mathcal{N}_0$. Then the
following min-max problem
\begin{equation}\label{equ.robust}
\min_{\mathbf{w},b}\sup_{(\boldsymbol{\delta}_1,\cdots,
\boldsymbol{\delta_m})\in
\mathcal{N}}\left\{r(\mathbf{w},b)+\sum_{i=1}^m \max\big[1-y_i
(\langle\mathbf{w},\mathbf{x}_i-\boldsymbol{\delta}_i\rangle+b),\,
0\big]\right\}
\end{equation} is equivalent to the following optimization problem
on $\mathbf{w},b,\boldsymbol{\xi}$:
\begin{equation}\label{equ.nonsep111}
\begin{split}
\min: \quad& r(\mathbf{w},b)+\sup_{\boldsymbol{\delta}\in \mathcal{N}_0} (\mathbf{w}^\top \boldsymbol{\delta})+\sum_{i=1}^m \xi_i,\\
\st: \quad &\xi_i \geq
1-[y_i(\langle\mathbf{w},\,\mathbf{x}_i\rangle+b)], \quad i=1,\dots, m;\\
&\xi_i\geq 0, \quad i=1, \dots, m.
\end{split}
\end{equation}
Furthermore, the minimization of Problem~(\ref{equ.nonsep111}) is
attainable when $r(\cdot,\cdot)$ is lower semi-continuous.
\end{theorem}
\begin{proof} Define:
\begin{eqnarray*}
v(\mathbf{w},b) &\triangleq&
\sup_{\boldsymbol{\delta}\in\mathcal{N}_0}(\mathbf{w}^\top
\boldsymbol{\delta})+\sum_{i=1}^m
\max\big[1-y_i(\langle\mathbf{w},\mathbf{x}_i\rangle+b),\, 0\big].
\end{eqnarray*}
Recall that $\mathcal{N}^-\subseteq \mathcal{N}\subseteq{N}^+$ by
definition. Hence, fixing any $(\hat{\mathbf{w}}, \hat{b})\in
\mathbb{R}^{n+1}$, the following inequalities hold:
\begin{equation*}
\begin{split}
&\sup_{(\boldsymbol{\delta}_1,\cdots, \boldsymbol{\delta_m})\in
\mathcal{N}^-}\sum_{i=1}^m \max\big[1-y_i
(\langle\hat{\mathbf{w}},\mathbf{x}_i-\boldsymbol{\delta}_i\rangle+\hat{b}),\,
0\big]\\
\leq&\sup_{(\boldsymbol{\delta}_1,\cdots, \boldsymbol{\delta_m})\in
\mathcal{N}}\sum_{i=1}^m \max\big[1-y_i
(\langle\hat{\mathbf{w}},\mathbf{x}_i-\boldsymbol{\delta}_i\rangle+\hat{b}),\,
0\big]\\
\leq&\sup_{(\boldsymbol{\delta}_1,\cdots, \boldsymbol{\delta_m})\in
\mathcal{N}^+}\sum_{i=1}^m \max\big[1-y_i
(\langle\hat{\mathbf{w}},\mathbf{x}_i-\boldsymbol{\delta}_i\rangle+\hat{b}),\,
0\big].
\end{split}
\end{equation*}
To prove the theorem, we first show that
$v(\hat{\mathbf{w}},\hat{b})$ is no larger than the leftmost
expression and then show $v(\hat{\mathbf{w}},\hat{b})$ is no smaller
than the rightmost expression.

\noindent Step 1: We prove that
\begin{equation}\label{equ.corprostep1}
v(\hat{\mathbf{w}},\hat{b}) \leq\sup_{(\boldsymbol{\delta}_1,\cdots,
\boldsymbol{\delta_m})\in \mathcal{N}^-}\sum_{i=1}^m \max\big[1-y_i
(\langle\hat{\mathbf{w}},\mathbf{x}_i-\boldsymbol{\delta}_i\rangle+\hat{b}),\,
0\big].
\end{equation}
Since the samples $\{\mathbf{x}_i,\, y_i\}_{i=1}^m$ are not
separable, there exists $t\in [1:m]$ such that
\begin{equation}
\label{equ.nonseparable}y_t(\langle\hat{\mathbf{w}},\mathbf{x}_t\rangle+\hat{b})<0.
\end{equation}
Hence,
\begin{eqnarray*}
&&\sup_{(\boldsymbol{\delta}_1,\cdots, \boldsymbol{\delta_m})\in
\mathcal{N}^-_t}\sum_{i=1}^m \max\big[1-y_i
(\langle\hat{\mathbf{w}},\mathbf{x}_i-\boldsymbol{\delta}_i\rangle+\hat{b}),\,
0\big]\\
&&=\sum_{i\not=t}\max\big[1-y_i
(\langle\hat{\mathbf{w}},\mathbf{x}_i\rangle+\hat{b}),\, 0\big]+
\sup_{\boldsymbol{\delta}_t\in \mathcal{N}_0}\max\big[1-y_t
(\langle\hat{\mathbf{w}},\mathbf{x}_t-\boldsymbol{\delta}_t\rangle+\hat{b}),\,
0\big]\\
&&=\sum_{i\not=t}\max\big[1-y_i
(\langle\hat{\mathbf{w}},\mathbf{x}_i\rangle+\hat{b}),\, 0\big]+
\max\big[1-y_t
(\langle\hat{\mathbf{w}},\mathbf{x}_t\rangle+\hat{b})+
\sup_{\boldsymbol{\delta}_t\in
\mathcal{N}_0}(y_t\hat{\mathbf{w}}^\top\boldsymbol{\delta}_t),\,
0\big]\\
&& =\sum_{i\not=t}\max\big[1-y_i
(\langle\hat{\mathbf{w}},\mathbf{x}_i\rangle+\hat{b}),\, 0\big]+
 \max\big[1-y_t
(\langle\hat{\mathbf{w}},\mathbf{x}_t\rangle+\hat{b}),\, 0\big]+
\sup_{\boldsymbol{\delta}_t\in
\mathcal{N}_0}(y_t\hat{\mathbf{w}}^\top\boldsymbol{\delta}_t)\\
&& =
\sup_{\boldsymbol{\delta}\in\mathcal{N}_0}(\hat{\mathbf{w}}^\top
\boldsymbol{\delta})+\sum_{i=1}^m
\max\big[1-y_i(\langle\hat{\mathbf{w}},\mathbf{x}_i\rangle+\hat{b}),
0\big]   =v(\hat{\mathbf{w}},\hat{b}).
\end{eqnarray*}
The third equality holds because of
Inequality~(\ref{equ.nonseparable}) and
$\sup_{\boldsymbol{\delta}_t\in
\mathcal{N}_0}(y_t\hat{\mathbf{w}}^\top\boldsymbol{\delta}_t)$ being
non-negative (recall $\mathbf{0}\in \mathcal{N}_0$). Since
$\mathcal{N}^-_t \subseteq \mathcal{N}^-$,
Inequality~(\ref{equ.corprostep1}) follows.

Step 2: Next we prove that
\begin{equation}\label{equ.nosproofstep2}
\sup_{(\boldsymbol{\delta}_1,\cdots, \boldsymbol{\delta_m})\in
\mathcal{N}^+}\sum_{i=1}^m \max\big[1-y_i
(\langle\hat{\mathbf{w}},\mathbf{x}_i-\boldsymbol{\delta}_i\rangle+\hat{b}),\,
0\big]\leq v(\hat{\mathbf{w}},\hat{b}).
\end{equation}
Notice that by the definition of $\mathcal{N}^+$ we have
\begin{equation}\label{equ.proofstep2-2}\begin{split}
&\sup_{(\boldsymbol{\delta}_1,\cdots, \boldsymbol{\delta_m})\in
\mathcal{N}^+}\sum_{i=1}^m \max\big[1-y_i
(\langle\hat{\mathbf{w}},\mathbf{x}_i-\boldsymbol{\delta}_i\rangle+\hat{b}),\,
0\big]\\
=&\sup_{\sum_{i=1}^m \alpha_i=1;\, \alpha_i\geq 0;\,
\hat{\boldsymbol{\delta}}_i\in \mathcal{N}_0}\sum_{i=1}^m
\max\big[1-y_i
(\langle\hat{\mathbf{w}},\mathbf{x}_i-\alpha_i{\hat{\boldsymbol{\delta}}}_i\rangle+\hat{b}),\,
0\big]\\
=&\sup_{\sum_{i=1}^m \alpha_i=1;\, \alpha_i\geq 0;}\sum_{i=1}^m
\max\big[\sup_{\hat{\boldsymbol{\delta}}_i\in
\mathcal{N}_0}\big(1-y_i
(\langle\hat{\mathbf{w}},\mathbf{x}_i-\alpha_i{\hat{\boldsymbol{\delta}}}_i\rangle+\hat{b})\big),\,
0\big].\end{split}\end{equation}

Now, for any $i\in [1:m]$, the following holds,
\begin{equation*}
\begin{split} &\max\big[\sup_{\hat{\boldsymbol{\delta}}_i\in
\mathcal{N}_0}\big(1-y_i
(\langle\hat{\mathbf{w}},\,\mathbf{x}_i-\alpha_i{\hat{\boldsymbol{\delta}}}_i\rangle+\hat{b})\big),\,
0\big]\\=&\max\big[1-y_i(\langle\hat{\mathbf{w}},\mathbf{x}_i\rangle+\hat{b})+\alpha_i\sup_{\hat{\boldsymbol{\delta}}_i\in
\mathcal{N}_0}(\hat{\mathbf{w}}^\top{\hat{\boldsymbol{\delta}}}_i),\,
0\big]
\\ \leq& \max\big[1-y_i
(\langle\hat{\mathbf{w}},\mathbf{x}_i\rangle+\hat{b}),\,
0\big]+\alpha_i\sup_{\hat{\boldsymbol{\delta}}_i\in
\mathcal{N}_0}(\hat{\mathbf{w}}^\top{\hat{\boldsymbol{\delta}}}_i).
\end{split}
\end{equation*}
Therefore, Equation~(\ref{equ.proofstep2-2}) is upper bounded by
\begin{eqnarray*}
&& \sum_{i=1}^m \max\big[1-y_i
(\langle\hat{\mathbf{w}},\mathbf{x}_i\rangle+\hat{b}),\, 0\big] +
 \sup_{\sum_{i=1}^m \alpha_i=1;\, \alpha_i\geq
0;}\sum_{i=1}^m\alpha_i\sup_{\hat{\boldsymbol{\delta}}_i\in
\mathcal{N}_0}(\hat{\mathbf{w}}^\top{\hat{\boldsymbol{\delta}}}_i) \\
&&=\sup_{\boldsymbol{\delta}\in\mathcal{N}_0}(\hat{\mathbf{w}}^\top\boldsymbol{\delta})+\sum_{i=1}^m
\max\big[1-y_i(\langle\hat{\mathbf{w}},\mathbf{x}_i\rangle+\hat{b}),
0\big]  =v(\hat{\mathbf{w}},\hat{b}),
\end{eqnarray*}
hence Inequality~(\ref{equ.nosproofstep2}) holds.

Step 3: Combining the two steps and adding $r(\mathbf{w},b)$ on both
sides leads to: $\forall(\mathbf{w},b)\in \mathbb{R}^{n+1}$,
\begin{equation*}
 \sup_{(\boldsymbol{\delta}_1,\cdots, \boldsymbol{\delta_m})\in
\mathcal{N}}\sum_{i=1}^m \max\big[1-y_i
(\langle\mathbf{w},\mathbf{x}_i-\boldsymbol{\delta}_i\rangle+b),\,
0\big] +r(\mathbf{w},b)  = v(\mathbf{w},b)+r(\mathbf{w,b}).
\end{equation*}
Taking the infimum on both sides establishes the equivalence of
Problem~(\ref{equ.robust}) and Problem~(\ref{equ.nonsep111}).
Observe that $\sup_{\boldsymbol{\delta}\in\mathcal{N}_0}
\mathbf{w}^\top \boldsymbol{\delta}$ is a supremum over a class of
affine functions, and hence is lower semi-continuous. Therefore
$v(\cdot,\cdot)$ is also lower semi-continuous. Thus the minimum can
be achieved for Problem~(\ref{equ.nonsep111}), and
Problem~(\ref{equ.robust}) by equivalence, when $r(\cdot)$ is lower
semi-continuous.\end{proof}
\oldversion{\noindent
 This theorem reveals the main difference in conservatism between
the constraint-wise uncertainty in (\ref{eq:naiverobust}) and our
formulation in (\ref{equ.robustclassifierformulation}). Consider
both formulations with the same uncertainty set, ${\cal N} =
\{(\boldsymbol{\delta}_1,\cdots,\boldsymbol{\delta}_m)|\sum_{i=1}^m
\|\boldsymbol{\delta}_i\| \leq c\}$. The corresponding atomic set of
${\cal N}$ is ${\cal N}_0 = \{\|\mathbf{\delta}\| \leq c\}$, but the
atomic set for ${\cal N}_{\mbox{\tiny{box}}}$ is $m {\cal N}_0$
since ${\cal N}_{\mbox{\tiny{box}}}={\cal N}_0\times \cdots\times
{\cal N}_0$ includes disturbances with a magnitude as large as $mc$.
Therefore the latter (recall (\ref{eq:boxrobust})) is equivalent to
a regularization coefficient of the form $mc$ that is {\it linked to the
number of training samples}, and will therefore be overly
conservative.}
\newversion{This theorem reveals the main difference between Formulation~(\ref{eq:boxrobust}) and our
formulation in~(\ref{equ.robustclassifierformulation}). Consider a
Sublinear Aggregated Uncertainty set ${\cal N} =
\{(\boldsymbol{\delta}_1,\cdots,\boldsymbol{\delta}_m)|\sum_{i=1}^m
\|\boldsymbol{\delta}_i\| \leq c\}$. The smallest box-type
uncertainty set containing $\cal{N}$ includes disturbances with norm
sum up to $mc$. Therefore, it leads to a regularization coefficient
as large as $mc$  that is linked to the number of training samples,
and will therefore be overly conservative.}

An immediate corollary is that a special case of our robust
formulation is equivalent to the norm-regularized SVM setup:
\begin{corollary}\label{cor.equiv}
 Let $\mathcal{T} \triangleq \Big\{(\boldsymbol{\delta}_1,\cdots
\boldsymbol{\delta}_m)| \sum_{i=1}^m \|\boldsymbol{\delta}_i\|^*\leq
c\Big\}$.
If the training sample $\{\mathbf{x}_i,y_i\}_{i=1}^m$ are
non-separable, then the following two optimization problems on
$(\mathbf{w}, b)$ are equivalent\footnote{The optimization
equivalence for the linear case was observed independently by
\citet{Bertsimas08}.}
\begin{eqnarray}
\min: && \quad
\max_{(\boldsymbol{\delta}_1,\cdots,\boldsymbol{\delta}_m) \in
\mathcal{T}} \sum_{i=1}^m \max\big[1-y_i
\big(\langle\mathbf{w},\,\mathbf{x}_i-\boldsymbol{\delta}_i\rangle+b\big),0\big],\\
\min: && \quad  c\|\mathbf{w}\|+ \sum_{i=1}^m \max\big[1-y_i
\big(\langle\mathbf{w},\,\mathbf{x}_i\rangle+b\big), 0\big].
\label{equ.spe.euiv}
\end{eqnarray}
\end{corollary}
\begin{proof} Let $\mathcal{N}_0$ be the dual-norm ball $\{\boldsymbol{\delta}|\|\boldsymbol{\delta}\|^*\leq
c\}$  and $r(\mathbf{w},b)\equiv 0$. Then
$\sup_{\|\boldsymbol{\delta}\|^*\leq c}
(\mathbf{w}^\top\boldsymbol{\delta})=c\|\mathbf{w}\|$. The corollary
follows from Theorem~\ref{thm.robustequiv}. Notice indeed the
equivalence holds for any $\mathbf{w}$ and $b$.
\end{proof}
This corollary explains the widely known fact that the regularized
classifier tends to be more robust. Specifically, it explains the
observation that when the disturbance is noise-like and neutral
rather than adversarial, a norm-regularized classifier (without any
robustness requirement) has a performance often superior to a {\em
box-typed} robust classifier \citep[see][]{Trafalis07}. On the other
hand, this observation also suggests that the appropriate way to
regularize should come from a disturbance-robustness perspective.
The above equivalence implies that standard regularization
essentially assumes that the disturbance is spherical; if this is
not true, robustness may yield a better regularization-like
algorithm. To find a more effective regularization term, a closer
investigation of the data variation is desirable, e.g., by examining
the variation of the data and solving the corresponding robust
classification problem. For example, one way to regularize is by
splitting the given training samples into two subsets with equal
number of elements, and treating one as a disturbed copy of the
other. By analyzing the direction of the disturbance and the
magnitude of the total variation, one can choose the proper norm to
use, and a suitable tradeoff parameter.
\section{Probabilistic Interpretations}\label{sss.probinter}
Although Problem~(\ref{equ.robust}) is formulated without any
probabilistic assumptions, in this section, we briefly explain two
approaches to construct the uncertainty set and equivalently tune
the regularization parameter $c$ based on probabilistic information.

The first approach is to use Problem~(\ref{equ.robust}) to
approximate an upper bound for a chance-constrained classifier.
Suppose the disturbance $(\boldsymbol{\delta}_1^r,\cdots
\boldsymbol{\delta}_m^r)$ follows a joint probability measure $\mu$.
Then the chance-constrained classifier is given by the following
minimization problem given a confidence level $\eta\in [0,\,1]$,
\begin{eqnarray}\label{equ.chanceconst}
\min_{\mathbf{w}, b, l}: && l \nonumber \\
\st:  && \mu\Big\{\sum_{i=1}^m
\max\big[1-y_i(\langle\mathbf{w},\,\mathbf{x}_i-\boldsymbol{\delta}_i^r\rangle+b),0\big]\leq
l\Big\}  \geq 1-\eta.
\end{eqnarray}
The formulations in \cite{Shivaswamy06}, \cite{Lanckriet02} and
\cite{Bhattacharyya04} assume uncorrelated noise and require all
constraints to be satisfied with high probability {\em
simultaneously}. They find a vector $[\xi_1, \cdots, \xi_m]^\top$
where each $\xi_i$ is the $\eta$-quantile of the hinge-loss for
sample $\mathbf{x}_i^r$. In contrast, our formulation above
minimizes the $\eta$-quantile of the average (or equivalently the
sum of) empirical error. When controlling this average quantity is
of more interest, the box-type noise formulation will be overly
conservative.

Problem~(\ref{equ.chanceconst}) is generally intractable. However,
we can approximate it as follows. Let
$$
c^*\triangleq\inf\{\alpha|\mu(\sum_i\|\boldsymbol{\delta}_i\|^* \leq
\alpha)\geq 1-\eta\}.
$$
Notice that $c^*$ is easily simulated given $\mu$. Then for any
$(\mathbf{w}, b)$, with probability no less than  $1-\eta$, the
following holds,
\begin{equation*}\begin{split}
&\sum_{i=1}^m
\max\big[1-y_i(\langle\mathbf{w},\,\mathbf{x}_i-\boldsymbol{\delta}^r_i\rangle+b),0\big]\\
\leq& \max_{\sum_i{\|\boldsymbol{\delta}_i\|^*\leq c^*}}
\sum_{i=1}^m \max\big[1-y_i
(\langle\mathbf{w},\,\mathbf{x}_i-\boldsymbol{\delta}_i\rangle+b),0\big].
\end{split}
\end{equation*}
Thus (\ref{equ.chanceconst}) is upper bounded by
(\ref{equ.spe.euiv}) with $c=c^*$. This gives an additional
probabilistic robustness property of the standard regularized
classifier. Notice that following a similar approach but with the
constraint-wise robust setup, i.e., the box uncertainty set, would
lead to considerably more pessimistic approximations of the chance
constraint.

The second approach considers a Bayesian setup. Suppose the total
disturbance $c^r\triangleq
\sum_{i=1}^m\|\boldsymbol{\delta}_i^r\|^*$ follows a prior
distribution $\rho(\cdot)$. This can model
 for example the case that the training
sample set is  a mixture of several data sets where the disturbance
magnitude of each set is known. Such a setup leads to the following
classifier which minimizes the Bayesian (robust) error:
\begin{equation}\label{equ.bayesian}\min_{\mathbf{w}, b}:  \quad\int\,\Big\{
\max_{\sum \|\boldsymbol{\delta}_i\|^*\leq c} \sum_{i=1}^m
\max\big[1-y_i
\big(\langle\mathbf{w},\,\mathbf{x}_i-\boldsymbol{\delta}_i\rangle+b\big),0\big]\Big\}
 \,d\rho(c).
\end{equation}
By Corollary~\ref{cor.equiv}, the Bayesian
classifier~(\ref{equ.bayesian}) is equivalent to
\begin{equation*}\min_{\mathbf{w},b}: \quad  \int \Big\{c\|\mathbf{w}\|+ \sum_{i=1}^m \max\big[1-y_i
\big(\langle\mathbf{w},\,\mathbf{x}_i\rangle+b\big), 0\big]\Big\}\,
d\rho(c),
\end{equation*}
which can be further simplified as
\begin{equation*}\min_{\mathbf{w},b}: \quad  \overline{c}\|\mathbf{w}\|+ \sum_{i=1}^m \max\big[1-y_i
\big(\langle\mathbf{w},\,\mathbf{x}_i\rangle+b\big), 0\big],
\end{equation*}
where $\overline{c}\triangleq \int c\,d\rho(c)$. This thus provides
us a justifiable parameter tuning method different from cross
validation: simply using the expected value of $c^r$. We note that
it is the equivalence of Corollary~\ref{cor.equiv} that makes this
possible, since it is difficult to imagine a setting where one would have
a prior on regularization coefficients.
\section{Kernelization}\label{sec.kernelization}
The previous results can be easily generalized to the kernelized
setting, which we discuss in detail in this section. In
particular, similar to the linear classification case, we give a new
interpretation of the standard kernelized SVM as the min-max
empirical hinge-loss solution, where the disturbance is assumed to lie in
the feature space. We then relate this to the (more intuitively
appealing) setup where the disturbance lies in the sample space.
We use this relationship in Section \ref{sss.consistency} to prove
a consistency result for kernelized SVMs.

The kernelized SVM formulation considers a linear classifier in the
feature space $\mathcal{H}$, a Hilbert space containing the range of
some feature mapping $\Phi(\cdot)$. The standard formulation is as
follows,
\begin{eqnarray*}
\min_{\mathbf{w},b}: && r(\mathbf{w},b) +\sum_{i=1}^m \xi_i \\
\st: &&  \xi_i \geq \big[1-y_i(\langle\mathbf{w}, \Phi(\mathbf{x}_i)\rangle+b)], \\
&& \xi_i \geq 0\,.
\end{eqnarray*}
It has been proved in \citet{Scholkopf02} that if we take $f(\langle
\mathbf{w}, \mathbf{w}\rangle)$ for some increasing function
$f(\cdot)$ as the regularization term $r(\mathbf{w}, b)$, then the
optimal solution has a representation $\mathbf{w}^*=\sum_{i=1}^m
\alpha_i \Phi(\mathbf{x}_i)$, which can further be solved without
knowing explicitly the feature mapping, but by evaluating a kernel
function $k(\mathbf{x}, \mathbf{x}')\triangleq \langle
\Phi(\mathbf{x}),\, \Phi(\mathbf{x}')\rangle$ only. This is the
well-known ``kernel trick''.

The definitions of Atomic Uncertainty Set and Sublinear Aggregated
Uncertainty Set in the feature space are identical to
Definition~\ref{def.concavecorrelate1}
and~\ref{def.concavecorrelate2}, with $\mathbb{R}^n$ replaced by
$\mathcal{H}$. The following theorem is a feature-space counterpart
of Theorem~\ref{thm.robustequiv}. The proof follows from a similar
argument to Theorem~\ref{thm.robustequiv}, i.e., for any fixed
$(\mathbf{w}, b)$ the worst-case empirical error equals the
empirical error plus a penalty term
$\sup_{\boldsymbol{\delta}\in\mathcal{N}_0}\big(\langle\mathbf{w},\,\boldsymbol{\delta}\rangle\big)$,
and hence the details are omitted.
\begin{theorem}
Assume $\{\Phi(\mathbf{x}_i), y_i\}_{i=1}^m$ are not linearly
separable, $r(\cdot):\mathcal{H}\times\mathbb{R}\rightarrow
\mathbb{R}$ is an arbitrary function,  $\mathcal{N}\subseteq
\mathcal{H}^m$ is a Sublinear Aggregated Uncertainty set with
corresponding atomic uncertainty set $\mathcal{N}_0\subseteq
\mathcal{H}$. Then the following min-max problem
\begin{equation}\label{equ.robust_kernel}
\min_{\mathbf{w},b}\sup_{(\boldsymbol{\delta}_1,\cdots,
\boldsymbol{\delta_m})\in
\mathcal{N}}\left\{r(\mathbf{w},b)+\sum_{i=1}^m \max\big[1-y_i
(\langle\mathbf{w},\Phi(\mathbf{x}_i)-\boldsymbol{\delta}_i\rangle+b),\,
0\big]\right\}
\end{equation} is equivalent to
\begin{equation}\label{equ.nonsep111_kernel}
\begin{split}
\min: \quad& r(\mathbf{w},b)+\sup_{\boldsymbol{\delta}\in \mathcal{N}_0} (\langle \mathbf{w},\, \boldsymbol{\delta}\rangle)+\sum_{i=1}^m \xi_i,\\
\st: \quad &\xi_i \geq
1-y_i\big(\langle\mathbf{w},\,\Phi(\mathbf{x}_i)\rangle+b\big), \quad i=1,\cdots, m;\\
&\xi_i\geq 0, \quad i=1, \cdots, m.
\end{split}
\end{equation}
Furthermore, the minimization of
Problem~(\ref{equ.nonsep111_kernel}) is attainable when
$r(\cdot,\cdot)$ is lower semi-continuous.
\end{theorem}

For some widely used feature mappings (e.g., RKHS of a Gaussian
kernel), $\{\Phi(\mathbf{x}_i), y_i\}_{i=1}^m$ are always separable.
In this case, the worst-case empirical error may not be equal to the
empirical error plus a penalty term
$\sup_{\boldsymbol{\delta}\in\mathcal{N}_0}\big(\langle\mathbf{w},\,\boldsymbol{\delta}\rangle\big)$.
However, it is easy to show that for any $(\mathbf{w},b)$, the
latter is an upper bound of the former.

The next corollary is the feature-space counterpart of
Corollary~\ref{cor.equiv}, where $\|\cdot\|_{\mathcal{H}}$ stands
for the RKHS norm, i.e., for $\mathbf{z}\in \mathcal{H}$,
$\|\mathbf{z}\|_{\mathcal{H}}=\sqrt{\langle\mathbf{z},\,\mathbf{z}\rangle}$.
Noticing that the RKHS norm is self dual, we find that the proof is
identical to that of Corollary~\ref{cor.equiv}, and hence omit it.
\begin{corollary}\label{cor.equiv.feature}
 Let $\mathcal{T}_{\mathcal{H}} \triangleq \Big\{(\boldsymbol{\delta}_1,\cdots
\boldsymbol{\delta}_m)| \sum_{i=1}^m
\|\boldsymbol{\delta}_i\|_{\mathcal{H}}\leq c \Big\}$.
If  $\{\Phi(\mathbf{x}_i),y_i\}_{i=1}^m$ are non-separable, then the
following two optimization problems on $(\mathbf{w}, b)$ are
equivalent
\begin{eqnarray}
\min: && \quad
\max_{(\boldsymbol{\delta}_1,\cdots,\boldsymbol{\delta}_m) \in
\mathcal{T}_{\mathcal{H}}} \sum_{i=1}^m \max\big[1-y_i
\big(\langle\mathbf{w},\,\Phi(\mathbf{x}_i)-\boldsymbol{\delta}_i\rangle+b\big),0\big],\\
\min: && \quad  c\|\mathbf{w}\|_{\mathcal{H}}+ \sum_{i=1}^m
\max\big[1-y_i
\big(\langle\mathbf{w},\,\Phi(\mathbf{x}_i)\rangle+b\big), 0\big].
\label{equ.spe.euiv1}
\end{eqnarray}
\end{corollary}
Equation~(\ref{equ.spe.euiv1}) is a variant form of the standard SVM
that has a squared RKHS norm regularization term, and it can be
shown that the two formulations are equivalent up to changing of
tradeoff parameter $c$, since both the empirical hinge-loss and the
RKHS norm are convex. Therefore, Corollary~\ref{cor.equiv.feature}
essentially means that the standard kernelized SVM is implicitly a
robust classifier (without regularization) with disturbance in the
feature-space, and the sum of the magnitude of the disturbance is bounded.

Disturbance in the feature-space is less intuitive than disturbance in
the sample space, and the next lemma relates these two different
notions.
\begin{lemma}\label{lem.noiserelevance}Suppose there exists $\mathcal{X}\subseteq \mathbb{R}^n$,
$\rho>0$, and a continuous non-decreasing function
$f:\mathbb{R}^+\rightarrow \mathbb{R}^+$ satisfying $f(0)=0$, such
that
\[k(\mathbf{x},
\mathbf{x})+k(\mathbf{x}',\mathbf{x}')-2k(\mathbf{x},
\mathbf{x}')\leq f(\|\mathbf{x}-\mathbf{x}'\|_2^2),\quad \forall
\mathbf{x}, \mathbf{x}' \in \mathcal{X},
\|\mathbf{x}-\mathbf{x}'\|_2\leq \rho\] then
\[\|\Phi(\hat{\mathbf{x}}+\boldsymbol{\delta})-\Phi(\hat{\mathbf{x}})\|_{\mathcal{H}}\leq \sqrt{f(\|\boldsymbol{\delta}\|_2^2)},\quad \forall \|\boldsymbol{\delta}\|_2\leq \rho,\,\, \hat{\mathbf{x}}, \hat{\mathbf{x}}+\boldsymbol{\delta}\in \mathcal{X}.\]
\end{lemma}
In the appendix, we prove a result that provides a tighter relationship between disturbance in the feature space and disturbance in the sample space, for RBF kernels.
\begin{proof}
Expanding the RKHS norm yields
\begin{equation*}
\begin{split}&\|\Phi(\hat{\mathbf{x}}+\boldsymbol{\delta})-\Phi(\hat{\mathbf{x}})\|_{\mathcal{H}}\\
=&\sqrt{\langle
\Phi(\hat{\mathbf{x}}+\boldsymbol{\delta})-\Phi(\hat{\mathbf{x}}),\,\Phi(\hat{\mathbf{x}}+\boldsymbol{\delta})-\Phi(\hat{\mathbf{x}})\rangle}\\
=&\sqrt{\langle
\Phi(\hat{\mathbf{x}}+\boldsymbol{\delta}),\,\Phi(\hat{\mathbf{x}}+\boldsymbol{\delta})\rangle
+\langle \Phi(\hat{\mathbf{x}}),\,\Phi(\hat{\mathbf{x}})\rangle
-2\langle
\Phi(\hat{\mathbf{x}}+\boldsymbol{\delta}),\,\Phi(\hat{\mathbf{x}})\rangle
}\\
=&\sqrt{k\big(
 \hat{\mathbf{x}}+\boldsymbol{\delta} ,\, \hat{\mathbf{x}}+\boldsymbol{\delta} \big)
+k\big( \hat{\mathbf{x}} ,\, \hat{\mathbf{x}} \big) -2k\big(
 \hat{\mathbf{x}}+\boldsymbol{\delta} ,\,\ \hat{\mathbf{x}} \big)
}\\
\leq&
\sqrt{f(\|\hat{\mathbf{x}}+\boldsymbol{\delta}-\hat{\mathbf{x}}\|_2^2)}
= \sqrt{f(\| \boldsymbol{\delta} \|_2^2)},
\end{split}
\end{equation*}
where the inequality follows from the assumption.
\end{proof}
Lemma~\ref{lem.noiserelevance} essentially says that under certain
conditions, robustness in the feature space is a stronger
requirement that robustness in the sample space. Therefore, a
classifier that achieves robustness in the feature space (the SVM
for example) also achieves robustness in the sample space. Notice
that the condition of Lemma~\ref{lem.noiserelevance} is rather
weak. In particular, it holds for any continuous $k(\cdot,\cdot)$
and bounded $\mathcal{X}$.

In the next section we consider a more foundational property of robustness in the sample
space: we show that a classifier that is robust in the sample space is asymptotically consistent.
As a consequence of this result for linear classifiers, the above results imply the consistency
for a broad class of kernelized SVMs.

\section{Consistency of Regularization}\label{sss.consistency}
In this section we explore a fundamental connection between
learning and robustness, by using robustness properties to re-prove the
statistical consistency of the linear classifier, and then the kernelized SVM.
Indeed, our proof mirrors the consistency proof found in \citep{Steinwart05}, with the
key difference that {\it we replace metric entropy, VC-dimension, and stability conditions used
there, with a robustness condition}.

Thus far we have considered the setup where the
training-samples are corrupted by certain set-inclusive
disturbances. We now turn to the standard statistical learning
setup, by assuming that all training samples and testing samples are
generated i.i.d.\ according to a (unknown) probability $\mathbb{P}$,
i.e., there does not exist explicit disturbance.

Let $\mathcal{X}\subseteq \mathbb{R}^n$ be bounded, and suppose the
training samples $(\mathbf{x}_i, y_i)_{i=1}^{\infty}$ are generated
i.i.d.\ according to an unknown distribution $\mathbb{P}$ supported
by $\mathcal{X}\times \{-1,\,+1\}$. The next theorem shows that our
robust classifier setup and equivalently regularized SVM asymptotically minimizes
an upper-bound of the expected classification error and
hinge loss.
\begin{theorem}\label{thm.consistencylinear}Denote $K\triangleq \max_{x\in
\mathcal{X}} \|x\|_2$. Then there exists a random sequence
$\{\gamma_{m,c}\}$ such that:
\begin{enumerate}
\item $\forall c > 0$, $\lim_{m \rightarrow \infty} \gamma_{m,c} = 0$ almost surely, and the convergence
is uniform in $\mathbb{P}$;
\item the following bounds on the Bayes loss and
the hinge loss hold uniformly for all $(\mb{w},b)$:
\begin{equation*}
\begin{split}
& \mathbb{E}_{(\mathbf{x}, y)\sim\mathbb{P}}(\mathbf{1}_{y\not=
sgn(\langle\mathbf{w},\,\mathbf{x}\rangle+b)})\leq
\gamma_{m,c}+c\|\mathbf{w}\|_2+ \frac{1}{m}\sum_{i=1}^m
\max\big[1-y_i(\langle\mathbf{w},\,\mathbf{x}_i\rangle+b), 0\big];\\
 &\mathbb{E}_{(\mathbf{x},
y)\sim\mathbb{P}}\big(\max(1-y(\langle\mathbf{w},\,\mathbf{x}\rangle+b),\,
0)\big)\leq\\&\quad\quad\quad\quad
\gamma_{m,c}(1+K\|\mathbf{w}\|_2+|b|)+c\|\mathbf{w}\|_2+
\frac{1}{m}\sum_{i=1}^m
\max\big[1-y_i(\langle\mathbf{w},\,\mathbf{x}_i\rangle+b), 0\big].
\end{split}
\end{equation*}
\end{enumerate}
\end{theorem}
\begin{proof} We briefly explain the basic idea of the proof before
going to the technical details. We consider the testing sample set
as a perturbed copy of the training sample set, and measure the
magnitude of the perturbation. For testing samples that have
``small'' perturbations, $c\|\mathbf{w}\|_2+ \frac{1}{m}\sum_{i=1}^m
\max\big[1-y_i(\langle\mathbf{w},\,\mathbf{x}_i\rangle+b), 0\big]$
 upper-bounds their total loss by Corollary~\ref{cor.equiv}. Therefore,
 we only need to show that the ratio of testing samples having
 ``large'' perturbations diminishes to prove the theorem.

Now we present the detailed proof. Given a $c>0$, we call a testing sample
$(\mathbf{x}', y')$ and a training sample $(\mathbf{x}, y)$ a {\em sample pair} if $y=y'$ and
$\|\mathbf{x}-\mathbf{x}'\|_2\leq c$. We say a set of training
samples and a set of testing samples form $l$ pairings if there
exist $l$ sample pairs with no data reused. Given $m$ training
samples and $m$ testing samples, we use $M_{m,c}$ to denote the
largest number of pairings. To prove this theorem, we need to
establish the following lemma.
\begin{lemma}\label{lem.pairnumber}Given a $c>0$, $M_{m,c}/m \rightarrow 1$ almost surely as $m\rightarrow
+\infty$, uniformly w.r.t. $\mathbb{P}$.
\end{lemma}
\begin{proof}We make a partition of $\mathcal{X}\times\{-1,\,+1\}=\bigcup_{t=1}^{T_c}\mathcal{X}_t $ such that
$\mathcal{X}_t$ either has the form
$[\alpha_1,\alpha_1+c/\sqrt{n})\times
[\alpha_2,\alpha_2+c/\sqrt{n})\cdots\times
[\alpha_n,\alpha_n+c/\sqrt{n})\times\{+1\}$ or
$[\alpha_1,\alpha_1+c/\sqrt{n})\times
[\alpha_2,\alpha_2+c/\sqrt{n})\cdots\times
[\alpha_n,\alpha_n+c/\sqrt{n})\times\{-1\}$ (recall $n$ is the
dimension of $\mathcal{X}$). That is, each partition is the
Cartesian product of a rectangular cell in $\mathcal{X}$ and a
singleton in $\{-1,\,+1\}$. Notice that if a training sample and a
testing sample fall into $\mathcal{X}_t$, they can form a pairing.

Let $N_t^{tr}$ and $N_{t}^{te}$ be the number of training samples
and testing samples falling in the $t^{th}$ set, respectively. Thus,
$(N_1^{tr},\cdots, N_{T_c}^{tr})$ and $(N_1^{te},\cdots,
N_{T_c}^{te})$ are multinomially distributed random vectors
following a same distribution. Notice that for a multinomially
distributed random vector $(N_1, \cdots, N_k)$ with parameter $m$
and $(p_1, \cdots, p_k)$, the following holds
\citep[Breteganolle-Huber-Carol inequality, see for example
Proposition A6.6 of][]{Vaart2000}. For any $\lambda > 0$,
\[\mathbb{P}\Big(\sum_{i=1}^k\big|N_i-mp_i\big|)\geq 2\sqrt{m}\lambda\Big)\leq
2^k\exp(-2\lambda^2).\]
Hence we have
\begin{eqnarray}\label{equ.mcconvergence}
&&\mathbb{P}\Big(\sum_{t=1}^{T_c} \big|N_t^{tr}-N_t^{te}\big|\geq
4\sqrt{m}\lambda\Big)\leq
2^{{T_c}+1}\exp(-2\lambda^2),\nonumber\\\Longrightarrow\quad &&
\mathbb{P}\Big(\frac{1}{m}\sum_{t=1}^{T_c} \big|N_t^{tr}-N_t^{te}\big|\geq
\lambda\Big)\leq 2^{{T_c}+1}\exp(\frac{-m\lambda^2}{8}),\nonumber\\
\Longrightarrow\quad && \mathbb{P}\Big(M_{m,c}/m\leq 1-\lambda\Big)\leq
2^{{T_c}+1}\exp(\frac{-m\lambda^2}{8}),
\end{eqnarray}
Observe that $\sum_{m=1}^{\infty}
2^{{T_c}+1}\exp(\frac{-m\lambda^2}{8}) < +\infty$, hence by the
Borel-Cantelli Lemma \citep[see for example][]{Durrett04}, with
probability one the event $\{M_{m,c}/m\leq 1-\lambda\}$ only occurs
finitely often as $m\rightarrow \infty$. That is, $\lim\inf_m
M_{m,c}/m \geq 1-\lambda$ almost surely. Since $\lambda$ can be
arbitrarily close to zero, $M_{m,c}/m \rightarrow 1$ almost surely.
Observe that this convergence is uniform in $\mathbb{P}$, since
$T_c$ only depends on $\mathcal{X}$.
\end{proof}
Now we proceed to prove the theorem. Given $m$ training samples and
$m$  testing samples with $M_{m,c}$ sample pairs, we notice that for
these paired samples, both the total testing error and the total
testing hinge-loss is upper bounded by
\begin{equation*}
\begin{split} &\max_{(\boldsymbol{\delta}_1,\cdots,\boldsymbol{\delta}_m) \in
\mathcal{N}_0 \times \cdots \times \mathcal{N}_0}
\sum_{i=1}^m\max\big[1-y_i
\big(\langle\mathbf{w},\,\mathbf{x}_i-\boldsymbol{\delta}_i\rangle+b\big),0\big]\\\leq&
cm\|\mathbf{w}\|_2+\sum_{i=1}^m
\max\big[1-y_i(\langle\mathbf{w},\,\mathbf{x}_i\rangle+b),\,0],
\end{split}\end{equation*}
where $\mathcal{N}_0 = \{\mb{\delta} \,|\, \|\mb{\delta}\| \leq c\}$.
Hence the total classification error of the $m$ testing samples can
be upper bounded by
\[(m-M_{m,c})+cm\|\mathbf{w}\|_2+\sum_{i=1}^m
\max\big[1-y_i(\langle\mathbf{w},\,\mathbf{x}_i\rangle+b),\,0],\]
and since \[\max_{\mathbf{x}\in \mathcal{X}} (1-y(\langle\mathbf{w},
\mathbf{x}\rangle))\leq \max_{\mathbf{x}\in \mathcal{X}}\Big\{
1+|b|+\sqrt{\langle \mathbf{x},\mathbf{x}\rangle\cdot \langle
\mathbf{w},\mathbf{w}\rangle}\Big\}=1+|b|+K\|\mathbf{w}\|_2,\]
the accumulated hinge-loss of the total $m$ testing samples is
upper bounded by
\[(m-M_{m,c})(1+K\|\mathbf{w}\|_2+|b|)+cm\|\mathbf{w}\|_2+\sum_{i=1}^m
\max\big[1-y_i(\langle\mathbf{w},\,\mathbf{x}_i\rangle+b),\,0].\]

Therefore, the average testing error  is  upper bounded by
\begin{equation}
\label{eq:avgtestinglossbound}
1-M_{m,c}/m+c\|\mathbf{w}\|_2+\frac{1}{m}\sum_{i=1}^n
\max\big[1-y_i(\langle\mathbf{w},\,\mathbf{x}_i\rangle+b),\,0],
\end{equation}
and the average hinge loss is upper bounded
by\[(1-M_{m,c}/m)(1+K\|\mathbf{w}\|_2+|b|)+c\|\mathbf{w}\|_2+
\frac{1}{m}\sum_{i=1}^m
\max\big[1-y_i(\langle\mathbf{w},\,\mathbf{x}_i\rangle+b), 0\big].\]
Let $\gamma_{m,c}=1-M_{m,c}/m$. The proof follows since
$M_{m,c}/m\rightarrow 1$ almost surely for any $c>0$. Notice by
Inequality~(\ref{equ.mcconvergence}) we have
\begin{equation}\label{equ.gammamcconverge}\mathbb{P}\Big(\gamma_{m,c}\geq \lambda\Big)\leq
\exp\Big({-m\lambda^2}/{8}+(T_c+1)\log2\Big),\end{equation}i.e., the
convergence is uniform in $\mathbb{P}$.

We have shown that the average testing error is upper bounded. The
final step is to show that this implies that in fact the random
variable given by the conditional expectation (conditioned on the
training sample) of the error is bounded almost surely as in the
statement of the theorem. To make things precise, consider a fixed
$m$, and let $\omega_1 \in \Omega_1$ and $\omega_2 \in \Omega_2$
generate the $m$ training samples and $m$ testing samples,
respectively, and for shorthand let $\mathcal{T}^m$ denote the
random variable of the first $m$ training samples. Let us denote the
probability measures for the training by $\rho_1$ and the testing
samples by $\rho_2$. By independence, the joint measure is given by
the product of these two. We rely on this property in what follows.
Now fix a $\lambda$ and a $c>0$. In our new notation,
Equation~(\ref{equ.gammamcconverge}) now reads:
\begin{eqnarray*}
\int_{\Omega_1}\int_{\Omega_2}
\mathbf{1}\big\{\gamma_{m,c}(\omega_1,\omega_2)\geq \lambda\big\}\,
d\rho_2(\omega_2) \, d\rho_1(\omega_1) &=&
\mathbb{P}\Big(\gamma_{m,c}(\omega_1,\omega_2)\geq \lambda\Big) \\
&\leq&
\exp\Big({-m\lambda^2}/{8}+(T_c+1)\log2\Big).
\end{eqnarray*}
We now bound $\mathbb{P}_{\omega_1}(\mathbb{E}_{\omega_2}[\gamma_{m,c}(\omega_1, \omega_2) \,|\, \mathcal{T}^m]>
\lambda)$, and then use Borel-Cantelli to show that this even can happen only
finitely often. We have:
\begin{eqnarray*}&&
\mathbb{P}_{\omega_1}(\mathbb{E}_{\omega_2}[\gamma_{m,c}(\omega_1,
\omega_2) \,|\, \mathcal{T}^m]> \lambda)\\&=&  \int_{\Omega_1}
\mathbf{1}\big\{\int_{\Omega_2} \gamma_{m,c}(\omega_1,\omega_2) \,
d\rho_2(\omega_2)>\lambda
\big\}\,d\rho_1(\omega_1)\\
&\leq& \int_{\Omega_1}\mathbf{1}\Big\{\big[
\int_{\Omega_2}\gamma_{m,c}(\omega_1,\omega_2) \mathbf{1}
(\gamma_{m,c}(\omega_1,\omega_2)\leq \lambda)\,
d\rho_2(\omega_2)+\\
&& \int_{\Omega_2}\gamma_{m,c}(\omega_1,\omega_2) \mathbf{1}
(\gamma_{m,c}(\omega_1,\omega_2)> \lambda)\, d\rho_2(\omega_2)
\big]\geq 2\lambda \Big\}d \rho_1(\omega_1)\\
&\leq& \int_{\Omega_1}\mathbf{1}\Big\{\big[ \int_{\Omega_2}\lambda
\mathbf{1} (\lambda(\omega_1,\omega_2)\leq \lambda)\,
d\rho_2(\omega_2)+ \\
&& \int_{\Omega_2} \mathbf{1}
(\gamma_{m,c}(\omega_1,\omega_2)> \lambda)\, d\rho_2(\omega_2)
\big]\geq 2\lambda \Big\}d \rho_1(\omega_1)\\
&\leq& \int_{\Omega_1}\mathbf{1}\Big\{\big[ \lambda + \int_{\Omega_2}
\mathbf{1} (\gamma_{m,c}(\omega_1,\omega_2)> \lambda)\,
d\rho_2(\omega_2)
\big]\geq 2\lambda \Big\}d \rho_1(\omega_1)\\
&=& \int_{\Omega_1}\mathbf{1}\Big\{ \int_{\Omega_2} \mathbf{1}
(\gamma_{m,c}(\omega_1,\omega_2)> \lambda)\, d\rho_2(\omega_2) \geq
\lambda \Big\}d \rho_1(\omega_1).
\end{eqnarray*}
Here, the first equality holds because training and testing samples are independent,
and hence the joint measure is the product of $\rho_1$ and $\rho_2$. The
second inequality holds because
$\gamma_{m,c}(\omega_1,\omega_2) \leq 1$ everywhere. Further notice that
\begin{equation*}
\begin{split}
&\int_{\Omega_1}\int_{\Omega_2}
\mathbf{1}\big\{\gamma_{m,c}(\omega_1,\omega_2)\geq \lambda\big\}\,
d\rho_2(\omega_2) \, d\rho_1(\omega_1) \\& \geq \int_{\Omega_1}
\lambda \mathbf{1} \Big\{ \int_{\Omega_2}
\mathbf{1}\big(\gamma_{m,c}(\omega_1,\omega_2)\geq \lambda\big) \,
d\rho(\omega_2) > \lambda \Big\}\, d\rho_1(\omega_1).
\end{split}
\end{equation*}
Thus we have
\[\mathbb{P}(\mathbb{E}_{\omega_2}(\gamma_{m,c}(\omega_1, \omega_2))>
\lambda) \leq \mathbb{P}\Big(\gamma_{m,c}(\omega_1,\omega_2)\geq
\lambda\Big)/\lambda\leq
\exp\Big({-m\lambda^2}/{8}+(T_c+1)\log2\Big)/\lambda.\] For any
$\lambda$ and $c$, summing up the right hand side over $m=1$ to
$\infty$ is finite, hence the theorem follows from the Borel-Cantelli
lemma.
\end{proof}

\begin{remark}{\rm We notice that, $M_m/m$ converges to $1$ almost surely even when $\mathcal{X}$ is not bounded. Indeed, to see
this, fix $\epsilon>0$, and let $\mathcal{X}'\subseteq \mathcal{X}$
be a bounded set such that $\mathbb{P}(\mathcal{X}')>1-\epsilon$.
Then, with probability one,
$$
\#(\mbox{unpaired samples in}\mathcal{X}')/m \rightarrow 0,
$$
by Lemma~\ref{lem.pairnumber}. In
addition,
\[\max\big(\#(\mbox{training samples not in }\mathcal{X}'),\,\#(\mbox{testing samples not in }\mathcal{X}')\big)/m\rightarrow \epsilon.\]
Notice that \begin{equation*}\begin{split}M_m&\geq
m-\#(\mbox{unpaired samples in
}\mathcal{X}')\\&\quad\quad-\max\big(\#(\mbox{training samples not
in }\mathcal{X}'),\,\#(\mbox{testing samples not in
}\mathcal{X}')\big).\end{split}\end{equation*}
Hence
$$
\lim_{m\rightarrow \infty}M_m/m \geq 1-\epsilon,
$$
almost surely.
Since $\epsilon$ is arbitrary, we have $M_m/m\rightarrow 1$ almost
surely.}
\end{remark}

Next, we prove an analog of Theorem \ref{thm.consistencylinear} for the kernelized
case, and then show that these two imply statistical consistency of linear and kernelized
SVMs. Again, let
$\mathcal{X}\subseteq \mathbb{R}^n$ be bounded, and suppose the
training samples $(\mathbf{x}_i, y_i)_{i=1}^{\infty}$ are generated
i.i.d.\ according to an unknown distribution $\mathbb{P}$ supported
on $\mathcal{X}\times \{-1,\,+1\}$.
\begin{theorem}\label{thm.consistencykernel}Denote
$K\triangleq\max_{\mathbf{x}\in \mathcal{X}} k(\mathbf{x},
\mathbf{x})$. Suppose there exists $\rho>0$ and a continuous
non-decreasing function $f:\mathbb{R}^+\rightarrow \mathbb{R}^+$
satisfying $f(0)=0$, such that:
\[k(\mathbf{x},
\mathbf{x})+k(\mathbf{x}',\mathbf{x}')-2k(\mathbf{x},
\mathbf{x}')\leq f(\|\mathbf{x}-\mathbf{x}'\|_2^2),\quad \forall
\mathbf{x}, \mathbf{x}' \in \mathcal{X},
\|\mathbf{x}-\mathbf{x}'\|_2\leq \rho.\]
Then there exists a random
sequence $\{\gamma_{m,c}\}$ such that,
\begin{enumerate}
\item $\forall c > 0$, $\lim_{m \rightarrow \infty} \gamma_{m,c} = 0$ almost
surely, and the convergence is uniform in $\mathbb{P}$; \item  the
following bounds on the Bayes loss and the hinge loss hold uniformly
for all $(\mb{w},b) \in \mathcal{H} \times \mathbb{R}$
\begin{equation*}
\begin{split}
&\mathbb{E}_{\mathbb{P}}(\mathbf{1}_{y\not=
sgn(\langle\mathbf{w},\,\Phi(\mathbf{x})\rangle+b)})\leq
\gamma_{m,c}+c\|\mathbf{w}\|_\mathcal{H}+ \frac{1}{m}\sum_{i=1}^m
\max\big[1-y_i(\langle\mathbf{w},\,\Phi(\mathbf{x}_i)\rangle+b),
0\big],\\
&\mathbb{E}_{(\mathbf{x},
y)\sim\mathbb{P}}\big(\max(1-y(\langle\mathbf{w},\,\Phi(\mathbf{x})\rangle+b),\,
0)\big)\leq\\
&\quad\quad\gamma_{m,c}(1+K\|\mathbf{w}\|_{\mathcal{H}}+|b|)+c\|\mathbf{w}\|_{\mathcal{H}}+
\frac{1}{m}\sum_{i=1}^m
\max\big[1-y_i(\langle\mathbf{w},\,\Phi(\mathbf{x}_i)\rangle+b),
0\big].
\end{split}
\end{equation*}
\end{enumerate}
\end{theorem}
\begin{proof} As in the proof of
Theorem~\ref{thm.consistencylinear}, we generate a set of $m$
testing samples and $m$ training samples, and then lower-bound the
number of samples that can form a {\em sample pair} in the
feature-space; that is, a pair consisting of a training sample $(\mathbf{x}, y)$
and a testing sample $(\mathbf{x}',y')$ such that $y=y'$ and
$\|\Phi(\mathbf{x})-\Phi(\mathbf{x}')\|_{\mathcal{H}}\leq c$.
In contrast to the finite-dimensional sample space, the
feature space may be infinite dimensional, and thus our decomposition
may have an infinite number of ``bricks.'' In this case, our multinomial
random variable argument used in the proof of Lemma~\ref{lem.pairnumber}
breaks down. Nevertheless, we are able to lower bound the number of
sample pairs in the feature space by the number of sample pairs in
the {\em sample space}.

Define $f^{-1}(\alpha)\triangleq \max\{\beta\geq0|f(\beta)\leq
\alpha\}$. Since $f(\cdot)$ is continuous, $f^{-1}(\alpha)>0$ for
any $\alpha>0$. Now notice that by Lemma~\ref{lem.noiserelevance},
if a testing sample $\mathbf{x}$ and a training sample $\mathbf{x}'$
belong to a ``brick'' with length of each side $\min(\rho/\sqrt{n},
\, f^{-1}(c^2)/\sqrt{n})$ in the {\em sample space} (see the proof
of Lemma~\ref{lem.pairnumber}),
$\|\Phi(\mathbf{x})-\Phi(\mathbf{x}')\|_{\mathcal{H}}\leq c$. Hence
the number of {\em sample pairs} in the feature space is lower
bounded by the number of pairs of samples that fall in the same
brick in the sample space. We can cover $\mathcal{X}$ with finitely
many (denoted as $T_c$) such bricks since $f^{-1}(c^2)>0$. Then, a
similar argument as in Lemma~\ref{lem.pairnumber} shows that the
ratio of samples that form pairs in a brick converges to $1$ as $m$
increases. Further notice that for $M$ paired samples, the total testing
error and hinge-loss are both upper-bounded by
\[cM\|\mathbf{w}\|_\mathcal{H}+\sum_{i=1}^M
\max\big[1-y_i(\langle\mathbf{w},\,\Phi(\mathbf{x}_i)\rangle+b),
0\big].\] The rest of the proof is identical to
Theorem~\ref{thm.consistencylinear}. In particular,
Inequality~(\ref{equ.gammamcconverge}) still holds.
\end{proof}
Notice that the condition in Theorem~\ref{thm.consistencykernel} is
satisfied by most widely used kernels, e.g., homogeneous polynominal
kernels, and Gaussian RBF. This condition requires that the feature
mapping is ``smooth'' and hence preserves ``locality'' of the disturbance,
i.e., small disturbance in the sample space guarantees the corresponding
disturbance in the feature space is also small.  It is easy to construct
non-smooth kernel functions which do not generalize well. For
example, consider the following kernel:
\[k(\mathbf{x}, \mathbf{x}')=\left\{\begin{array}{ll} 1 &
\mathbf{x}=\mathbf{x}';\\ 0&
\mathbf{x}\not=\mathbf{x}'.\end{array}\right.\] A standard RKHS
regularized SVM using this kernel leads to a decision function
\[\mathrm{sign}(\sum_{i=1}^m\alpha_ik(\mathbf{x}, \mathbf{x}_i)+b),\]
which equals $\mathrm{sign}(b)$ and provides no
meaningful prediction if the testing sample $\mathbf{x}$ is not one
of the training samples. Hence as $m$ increases, the testing error
remains as large as $50\%$ regardless of the tradeoff parameter used
in the algorithm, while the training error can be made arbitrarily small
by fine-tuning the parameter.
\subsubsection*{Convergence to Bayes Risk}
Next we relate the results of Theorem~\ref{thm.consistencylinear}
and Theorem \ref{thm.consistencykernel} to the standard consistency
notion, i.e., convergence to the Bayes Risk \citep{Steinwart05}. The key point
of interest in our proof is the use of a robustness condition in place of
a VC-dimension or stability condition used in \citep{Steinwart05}.
The proof in \citep{Steinwart05} has 4 main steps. They show: (i) there always
exists a minimizer to the expected regularized (kernel) hinge loss;
(ii) the expected  regularized hinge loss of the minimizer converges to
the expected hinge loss as the regularizer goes to zero; (iii) if a sequence of
functions asymptotically have optimal expected hinge loss, then they
also have optimal expected loss; and (iv) the expected hinge loss of the
minimizer of the regularized {\it training} hinge loss concentrates around the
empirical regularized hinge loss. In \citep{Steinwart05}, this final step, (iv), is accomplished
using concentration inequalities derived from VC-dimension considerations,
and stability considerations.

Instead, we use our robustness-based results of Theorem~\ref{thm.consistencylinear} and Theorem
\ref{thm.consistencykernel} to replace these approaches
(Lemmas 3.21 and 3.22 in \citep{Steinwart05}) in proving step (iv), and
thus to establish the main result.

Recall that a classifier is a rule that assigns to every training
set $T=\{\mathbf{x}_i, y_i\}_{i=1}^m$ a measurable function $f_T$.
The risk of a measurable function $f:\mathcal{X}\rightarrow
\mathbb{R}$ is defined as
\[\mathcal{R}_{\mathbb{P}}(f)\triangleq \mathbb{P}(\{\mathbf{x}, y:
\mathrm{sign}f(\mathbf{x})\not=y\}).\] The smallest achievable risk
\[\mathcal{R}_{\mathbb{P}}\triangleq \inf\{\mathcal{R}_{\mathbb{P}}(f)|f\,
\mbox{measurable}\}\] is called the {\em Bayes Risk} of
$\mathbb{P}$. A classifier is said to be strongly uniformly
consistent is for all distributions $P$ on $\mathcal{X}\times [-1,
+1]$, the following holds almost surely.
\[\lim_{m\rightarrow \infty} \mathcal{R}_{\mathbb{P}}(f_T)=\mathcal{R}_{\mathbb{P}}.\]

Without loss of generality, we only consider the kernel version.
Recall a definition from \citet{Steinwart05}.
\begin{definition}Let $C(\mathcal{X})$ be the set of all continuous
functions defined on $\mathcal{X}$. Consider the mapping $I:
\mathcal{H}\rightarrow C(\mathcal{X})$ defined by
$I\mathbf{w}\triangleq \langle \mathbf{w},\, \Phi(\cdot)\rangle$. If
$I$ has a dense image, we call the kernel {\em universal}.
\end{definition}
Roughly speaking, if a kernel is universal, it is rich enough to satisfy
the condition of step (ii) above.
\begin{theorem} If a kernel satisfies the condition of Theorem
\ref{thm.consistencykernel}, and is  universal, then the Kernel SVM
with $c\downarrow 0$ sufficiently slowly is strongly uniformly
consistent.
\end{theorem}
\begin{proof}
We first introduce some notation, largely following \citep{Steinwart05}.
For some probability measure $\mu$
and $(\mathbf{w},b)\in\mathcal{H}\times \mathbb{R}$,
\[R_{L, \mu}((\mathbf{w},b))\triangleq \mathbb{E}_{(\mathbf{x},y)\sim
\mu} \big\{\max(0, 1-y(\langle \mathbf{w}, \Phi(\mathbf{x})\rangle
+b))\big\},\] is the expected hinge-loss under probability $\mu$,
and \[R_{L, \mu}^c((\mathbf{w},b))\triangleq
c\|\mathbf{w}\|_\mathcal{H}+\mathbb{E}_{(\mathbf{x},y)\sim \mu}
\big\{\max(0, 1-y(\langle \mathbf{w}, \Phi(\mathbf{x})\rangle
+b))\big\}\] is the regularized expected hinge-loss. Hence $R_{L,
\mathbb{P}}(\cdot)$ and $R_{L, \mathbb{P}}^c(\cdot)$ are the
expected hinge-loss and regularized expected hinge-loss under the
generating probability $\mathbb{P}$. If $\mu$ is the empirical
distribution of $m$ samples, we write $R_{L,m}(\cdot)$  and
$R_{L,m}^c(\cdot)$ respectively. Notice $R_{L,m}^c(\cdot)$ is the
objective function of the SVM. Denote its solution by $f_{m,c}$, i.e.,
the classifier we get by running SVM with $m$ samples and parameter
$c$. Further denote by $f_{\mathbb{P},c}\in \mathcal{H}\times
\mathbb{R}$ the minimizer of $R_{L, \mathbb{P}}^c(\cdot)$. The
existence of such a minimizer is proved in Lemma 3.1 of
\citet{Steinwart05} (step (i)). Let
\[\mathcal{R}_{L,
\mathbb{P}}\triangleq\min_{f\,\rm{measurable}}\mathbb{E}_{\mathbf{x},y\sim
\mathbb{P}} \Big\{\max(1-y f(\mathbf{x}),\,0\big)\},\] i.e., the
smallest achievable hinge-loss for all measurable functions.

The main content of our proof is to use
Theorems~\ref{thm.consistencylinear} and \ref{thm.consistencykernel}
to prove step (iv) in \citet{Steinwart05}. In particular, we show:
if $c\downarrow 0$ ``slowly'', we have with probability one
\begin{equation}\label{equ.riskconsist}\lim_{m\rightarrow \infty} R_{L, \mathbb{P}}(f_{m,c})
=\mathcal{R}_{L, \mathbb{P}}.
\end{equation}
To prove Equation~(\ref{equ.riskconsist}), denote by $\mathbf{w}(f)$
and $b(f)$ as the weight part and offset part of any classifier $f$.
Next, we bound the magnitude of
$f_{m,c}$ by using $R_{L,m}^c(f_{m,c})\leq R_{L,m}^c(\mathbf{0},0)\leq
1$, which leads to
\begin{equation*}\|\mathbf{w}(f_{m,c})\|_{\mathcal{H}}\leq 1/c\end{equation*} and
\begin{equation*}|b(f_{m,c})|\leq 2+K\|\mathbf{w}(f_{m,c})\|_{\mathcal{H}}\leq
2+K/c.\end{equation*}

>From Theorem \ref{thm.consistencykernel} (note that the bound holds
uniformly for all $(\mathbf{w}, b)$), we have
\begin{eqnarray*}
R_{L,\mathbb{P}}(f_{m,c})&\leq&
\gamma_{m,c}[1+K\|\mathbf{w}(f_{m,c})\|_{\mathcal{H}}+|b|]+R_{L,m}^c(f_{m,c})\\
&\leq & \gamma_{m,c}[3+2K/c]+R_{L,m}^c(f_{m,c})\\
&\leq & \gamma_{m,c}[3+2K/c]+R_{L,m}^c(f_{\mathbb{P},c})\\
&=&\mathcal{R}_{L, \mathbb{P}}+\gamma_{m,c}[3+2K/c]
+\big\{R_{L,m}^c(f_{\mathbb{P},c})-R_{L,\mathbb{P}}^c(f_{\mathbb{P},c})\big\}+\big\{R_{L,
\mathbb{P}}^c(f_{\mathbb{P},c})-\mathcal{R}_{L, \mathbb{P}}\big\}\\
&=&\mathcal{R}_{L, \mathbb{P}}+\gamma_{m,c}[3+2K/c]
+\big\{R_{L,m}(f_{\mathbb{P},c})-R_{L,\mathbb{P}}(f_{\mathbb{P},c})\big\}+\big\{R_{L,
\mathbb{P}}^c(f_{\mathbb{P},c})-\mathcal{R}_{L, \mathbb{P}}\big\}.
\end{eqnarray*}
The last inequality holds because $f_{m,c}$ minimizes $R_{L,m}^c$.

It is known  \citep[Proposition 3.2]{Steinwart05} (step (ii)) that if the
kernel used is rich enough, i.e., universal, then
\[\lim_{c\rightarrow 0} R_{L, \mathbb{P}}^c(f_{\mathbb{P}},c)=\mathcal{R}_{L,\mathbb{P}}.\]
For fixed $c>0$, we have
\[\lim_{m\rightarrow
\infty}R_{L,m}(f_{\mathbb{P},c})=R_{L,\mathbb{P}}(f_{\mathbb{P},c}),\]
almost surely due to the strong law of large numbers (notice that
$f_{\mathbb{P},c}$ is a fixed classifier), and
$\gamma_{m,c}[3+2K/c]\rightarrow 0$ almost surely. Notice that neither
convergence rate depends on $\mathbb{P}$. Therefore, if
$c\downarrow 0$ sufficiently slowly,\footnote{For example, we can take
$\{c(m)\}$ be the smallest number satisfying $c(m)\geq m^{-1/8}$ and
$T_{c(m)}\leq m^{1/8}/\log2-1$.
Inequality~(\ref{equ.gammamcconverge}) thus leads to
$\sum_{m=1}^{\infty} P(\gamma_{m, c(m)}/c(m)\geq m^{1/4})\leq
+\infty$ which implies uniform convergence of
$\gamma_{m,c(m)}/c(m)$.}
 we have almost surely
\[\lim_{m\rightarrow \infty}R_{L,\mathbb{P}}(f_{m,c})\leq \mathcal{R}_{L,
\mathbb{P}}.\] Now, for any $m$ and $c$, we have
$R_{L,\mathbb{P}}(f_{m,c})\geq \mathcal{R}_{L, \mathbb{P}}$ by
definition. This implies that Equation~(\ref{equ.riskconsist}) holds
almost surely, thus giving us step (iv).

Finally, Proposition 3.3. of \citep{Steinwart05} shows step (iii),
namely, approximating hinge loss is sufficient to guarantee
approximation of the Bayes loss. Thus
Equation~(\ref{equ.riskconsist}) implies that the risk of function
$f_{m,c}$ converges to Bayes risk.\end{proof}

\section{Concluding Remarks}\label{sec.conclude}
This work considers the relationship between robust and regularized
SVM classification. In particular, we prove that the standard
norm-regularized SVM classifier is in fact the solution to a robust
classification setup, and thus known results about regularized
classifiers extend to robust classifiers. To the best of our
knowledge, this is the first explicit such link between
regularization and robustness in pattern classification. This  link
suggests that norm-based regularization essentially builds in a
robustness to sample noise whose probability level sets are
symmetric, and moreover have the structure of the unit ball with
respect to the  dual of the  regularizing norm. It would be
interesting to understand the performance gains possible when the
noise does not have such characteristics, and the robust setup is
used in place of regularization with appropriately defined
uncertainty set.

Based on the robustness interpretation of the regularization term,
we re-proved the consistency of SVMs without direct appeal to
notions of metric entropy, VC-dimension, or stability. Our proof
suggests that the ability to handle disturbance is crucial for an
algorithm to achieve good generalization ability. In particular, for
``smooth'' feature mappings, the robustness to disturbance in the
observation space is guaranteed and hence SVMs achieve consistency.
On the other-hand, certain ``non-smooth'' feature mappings fail to
be consistent simply because for such kernels the robustness in the
feature-space (guaranteed by the regularization process) does not
imply robustness in the observation space.

\acks{We thank the editor and three anonymous reviewers for
significantly improving the accessibility of this manuscript. We
also benefited from comments from participants in ITA 2008.}

\appendix
\section*{Appendix A.} In this appendix we show
that for RBF kernels, it is possible to relate robustness in the
feature space and robustness in the sample space more directly.
\begin{theorem}\label{thm.appendix} Suppose the Kernel function has
the form $k(\mathbf{x},\mathbf{x}')=f(\|\mathbf{x}-\mathbf{x}'\|)$,
with $f:\mathbb{R}^+\rightarrow \mathbb{R}$ a decreasing function.
Denote by $\mathcal{H}$ the RKHS space of $k(\cdot,\cdot)$ and
$\Phi(\cdot)$ the corresponding feature mapping. Then we have for
any $\mathbf{x}\in \mathbb{R}^n$, $\mathbf{w}\in \mathcal{H}$ and
 $c>0$,
\[\sup_{\|\boldsymbol{\delta}\|\leq c} \langle \mathbf{w},\, \Phi(\mathbf{x}-\boldsymbol{\delta})\rangle =\sup_{\|\boldsymbol{\delta}_\phi\|_{\mathcal{H}}\leq \sqrt{2 f(0)-2f(c)}}\langle \mathbf{w},\,\Phi(\mathbf{x})+\boldsymbol{\delta}_\phi \rangle.\]
\end{theorem}
\begin{proof}
We  show that the left-hand-side is not larger than the
right-hand-side, and vice versa.

First we show
\begin{equation}\label{equ.leftsmaller}
\sup_{\|\boldsymbol{\delta}\|\leq c} \langle \mathbf{w},\,
\Phi(\mathbf{x}-\boldsymbol{\delta})\rangle
\leq\sup_{\|\boldsymbol{\delta}_\phi\|_{\mathcal{H}}\leq \sqrt{2
f(0)-2f(c)}}\langle
\mathbf{w},\,\Phi(\mathbf{x})-\boldsymbol{\delta}_\phi
\rangle.\end{equation} We notice that for any
$\|\boldsymbol{\delta}\|\leq c$, we have
\begin{equation*}
\begin{split}
&\langle \mathbf{w},\, \Phi(\mathbf{x}-\boldsymbol{\delta})\rangle\\
=&\Big\langle \mathbf{w},\,
\Phi(\mathbf{x})+\big(\Phi(\mathbf{x}-\boldsymbol{\delta})-\Phi(\mathbf{x})\big)\Big\rangle\\
=&\langle \mathbf{w},\, \Phi(\mathbf{x}) \rangle  + \langle
\mathbf{w},\, \Phi(\mathbf{x}-\boldsymbol{\delta})-\Phi(\mathbf{x})
\rangle \\
\leq & \langle \mathbf{w},\, \Phi(\mathbf{x}) \rangle+
\|\mathbf{w}\|_{\mathcal{H}}\cdot\|\Phi(\mathbf{x}-\boldsymbol{\delta})-\Phi(\mathbf{x})\|_{\mathcal{H}}\\
\leq& \langle \mathbf{w},\, \Phi(\mathbf{x}) \rangle+
\|\mathbf{w}\|_{\mathcal{H}}\sqrt{2f(0)-2f(c)}\\
=& \sup_{\|\boldsymbol{\delta}_\phi\|_{\mathcal{H}}\leq \sqrt{2
f(0)-2f(c)}}\langle
\mathbf{w},\,\Phi(\mathbf{x})-\boldsymbol{\delta}_\phi\rangle.
\end{split}
\end{equation*}
Taking the supremum over $\boldsymbol{\delta}$
establishes Inequality~(\ref{equ.leftsmaller}).

Next, we show the opposite inequality,
\begin{equation}\label{equ.rightsmaller}
\sup_{\|\boldsymbol{\delta}\|\leq c} \langle \mathbf{w},\,
\Phi(\mathbf{x}-\boldsymbol{\delta})\rangle
\geq\sup_{\|\boldsymbol{\delta}_\phi\|_{\mathcal{H}}\leq \sqrt{2
f(0)-2f(c)}}\langle
\mathbf{w},\,\Phi(\mathbf{x})-\boldsymbol{\delta}_\phi
\rangle.\end{equation} If $f(c)=f(0)$, then
Inequality~\ref{equ.rightsmaller} holds trivially, hence we only
consider the case that $f(c)<f(0)$. Notice that the inner product
is a continuous function in $\mathcal{H}$, hence for any
$\epsilon>0$, there exists a $\boldsymbol{\delta}_{\phi}'$ such that
\[\langle
\mathbf{w},\,\Phi(\mathbf{x})-\boldsymbol{\delta}_\phi' \rangle
>\sup_{\|\boldsymbol{\delta}_\phi\|_{\mathcal{H}}\leq \sqrt{2
f(0)-2f(c)}}\langle
\mathbf{w},\,\Phi(\mathbf{x})-\boldsymbol{\delta}_\phi
\rangle-\epsilon;\quad
\|\boldsymbol{\delta}_{\phi}'\|_{\mathcal{H}}< \sqrt{2f(0)-2f(c)}.\]
Recall that the RKHS space is the completion of the feature mapping,
thus there exists a sequence of $\{\mathbf{x}'_i\}\in \mathbb{R}^n$
such that
\begin{equation}\label{equ.convergenceinappendix}\Phi(\mathbf{x}'_i)\rightarrow
\Phi(\mathbf{x})-\boldsymbol{\delta}_\phi',\end{equation} which is
equivalent to
\[\big(\Phi(\mathbf{x}'_i)-\Phi(\mathbf{x})\big)\rightarrow -\boldsymbol{\delta}_\phi'.\]
This leads to
\begin{equation*}
\begin{split}
&\lim_{i\rightarrow
\infty}\sqrt{2f(0)-2f(\|\mathbf{x}'_i-\mathbf{x}\|)}\\
=&\lim_{i\rightarrow
\infty}\|\Phi(\mathbf{x}'_i)-\Phi(\mathbf{x})\|_{\mathcal{H}}\\=&\|
\boldsymbol{\delta}_\phi'\|_{\mathcal{H}}<\sqrt{2f(0)-2f(c)}.
\end{split}
\end{equation*}
Since $f$ is decreasing, we conclude that
$\|\mathbf{x}'_i-\mathbf{x}\|\leq c$ holds except for a finite
number of $i$. By~(\ref{equ.convergenceinappendix}) we have
\[\langle \mathbf{w},\,\Phi(\mathbf{x}_i')\rangle \rightarrow \langle
\mathbf{w},\,\Phi(\mathbf{x})-\boldsymbol{\delta}_\phi' \rangle
>\sup_{\|\boldsymbol{\delta}_\phi\|_{\mathcal{H}}\leq \sqrt{2
f(0)-2f(c)}}\langle
\mathbf{w},\,\Phi(\mathbf{x})-\boldsymbol{\delta}_\phi
\rangle-\epsilon,\] which means
\[\sup_{\|\boldsymbol{\delta}\|\leq c} \langle \mathbf{w},\,
\Phi(\mathbf{x}-\boldsymbol{\delta})\rangle
\geq\sup_{\|\boldsymbol{\delta}_\phi\|_{\mathcal{H}}\leq \sqrt{2
f(0)-2f(c)}}\langle
\mathbf{w},\,\Phi(\mathbf{x})-\boldsymbol{\delta}_\phi
\rangle-\epsilon.\] Since $\epsilon$ is arbitrary, we establish
Inequality~(\ref{equ.rightsmaller}).

Combining Inequality~(\ref{equ.leftsmaller}) and
Inequality~(\ref{equ.rightsmaller}) proves the theorem.
\end{proof}

\bibliography{Phd1}

\begin{thebibliography}{43}
\providecommand{\natexlab}[1]{#1}
\providecommand{\url}[1]{\texttt{#1}}
\expandafter\ifx\csname urlstyle\endcsname\relax
  \providecommand{\doi}[1]{doi: #1}\else
  \providecommand{\doi}{doi: \begingroup \urlstyle{rm}\Url}\fi

\bibitem[Anthony and Bartlett(1999)]{Anthony99}
M.~Anthony and P.~Bartlett.
\newblock \emph{Neural Network Learning: Theoretical Foundations}.
\newblock Cambridge University Press, 1999.

\bibitem[Bartlett and Mendelson(2002)]{Bartlett02}
P.~Bartlett and S.~Mendelson.
\newblock Rademacher and {G}aussian complexities: Risk bounds and structural
  results.
\newblock \emph{Journal of Machine Learning Research}, 3:\penalty0 463--482,
  November 2002.

\bibitem[Bartlett et~al.(2005)Bartlett, Bousquet, and Mendelson]{Bartlett05}
P.~Bartlett, O.~Bousquet, and S.~Mendelson.
\newblock Local {R}ademacher complexity.
\newblock \emph{The Annals of Statistics}, 33\penalty0 (4):\penalty0
  1497--1537, 2005.

\bibitem[Ben-Tal and Nemirovski(1999)]{Ben-tal99}
A.~Ben-Tal and A.~Nemirovski.
\newblock Robust solutions of uncertain linear programs.
\newblock \emph{Operations Research Letters}, 25\penalty0 (1):\penalty0 1--13,
  August 1999.

\bibitem[Bennett and Mangasarian(1992)]{Bennett92}
K.~Bennett and O.~Mangasarian.
\newblock Robust linear programming discrimination of two linearly inseparable
  sets.
\newblock \emph{Optimization Methods and Software}, 1\penalty0 (1):\penalty0
  23--34, 1992.

\bibitem[Bertsimas and Fertis(2008)]{Bertsimas08}
D.~Bertsimas and A.~Fertis.
\newblock Personal Correspondence, March 2008.

\bibitem[Bertsimas and Sim(2004)]{Bertsimas04}
D.~Bertsimas and M.~Sim.
\newblock The price of robustness.
\newblock \emph{Operations Research}, 52\penalty0 (1):\penalty0 35--53, January
  2004.

\bibitem[Bhattacharyya(2004)]{Bhattacharyya04c}
C.~Bhattacharyya.
\newblock Robust classification of noisy data using second order cone
  programming approach.
\newblock In \emph{Proceedings International Conference on Intelligent Sensing
  and Information Processing}, pages 433--438, Chennai, India, 2004.

\bibitem[Bhattacharyya et~al.(2004{\natexlab{a}})Bhattacharyya, Grate, Jordan,
  {El Ghaoui}, and Mian]{Bhattacharyya04}
C.~Bhattacharyya, L.~Grate, M.~Jordan, L.~{El Ghaoui}, and I.~Mian.
\newblock Robust sparse hyperplane classifiers: Application to uncertain
  molecular profiling data.
\newblock \emph{Journal of Computational Biology}, 11\penalty0 (6):\penalty0
  1073--1089, 2004{\natexlab{a}}.

\bibitem[Bhattacharyya et~al.(2004{\natexlab{b}})Bhattacharyya, Pannagadatta,
  and Smola]{Bhattacharyya04b}
C.~Bhattacharyya, K.~Pannagadatta, and A.~Smola.
\newblock A second order cone programming formulation for classifying missing
  data.
\newblock In Lawrence~K. Saul, Yair Weiss, and {L\'{e}on} Bottou, editors,
  \emph{Advances in Neural Information Processing Systems (NIPS17)}, Cambridge,
  MA, 2004{\natexlab{b}}. MIT Press.

\bibitem[Bi and Zhang(2004)]{Bi04}
J.~Bi and T.~Zhang.
\newblock Support vector classification with input data uncertainty.
\newblock In Lawrence~K. Saul, Yair Weiss, and {L\'{e}on} Bottou, editors,
  \emph{Advances in Neural Information Processing Systems (NIPS17)}, Cambridge,
  MA, 2004. MIT Press.

\bibitem[Bishop(1995)]{Bishop95}
C.~Bishop.
\newblock Training with noise is equivalent to tikhonov regularization.
\newblock \emph{Neural Computation}, 7\penalty0 (1):\penalty0 108--116, 1995.
\newblock \doi{10.1162/neco.1995.7.1.108}.
\newblock URL
  \url{http://www.mitpressjournals.org/doi/abs/10.1162/neco.1995.7.1.108}.

\bibitem[Boser et~al.(1992)Boser, Guyon, and Vapnik]{Boser92}
P.~Boser, I.~Guyon, and V.~Vapnik.
\newblock A training algorithm for optimal margin classifiers.
\newblock In \emph{Proceedings of the Fifth Annual ACM Workshop on
  Computational Learning Theory}, pages 144--152, New York, NY, 1992.

\bibitem[Bousquet and Elisseeff(2002)]{Bousquet02}
O.~Bousquet and A.~Elisseeff.
\newblock Stability and generalization.
\newblock \emph{Journal of Machine Learning Research}, 2:\penalty0 499--526,
  2002.

\bibitem[Christmann and Steinwart(2004)]{ChristmannSteinwart04}
A.~Christmann and I.~Steinwart.
\newblock On robust properties of convex risk minimization methods for pattern
  recognition.
\newblock \emph{Journal of Machine Learning Research}, 5:\penalty0 1007--1034,
  2004.

\bibitem[Christmann and Steinwart(2007)]{ChristmannSteinwart07}
A.~Christmann and I.~Steinwart.
\newblock Consistency and robustness of kernel based regression.
\newblock \emph{Bernoulli}, 13\penalty0 (3):\penalty0 799--819, 2007.

\bibitem[Christmann and {Van Messem}(2008)]{ChristmannMessem08}
A.~Christmann and A.~{Van Messem}.
\newblock Bouligand derivatives and robustness of support vector machines.
\newblock \emph{Journal of Machine Learning Research}, 9:\penalty0 915--936,
  2008.

\bibitem[Cortes and Vapnik(1995)]{Cortes95}
C.~Cortes and V.~Vapnik.
\newblock Support vector networks.
\newblock \emph{Machine Learning}, 20:\penalty0 1--25, 1995.

\bibitem[Durrett(2004)]{Durrett04}
R.~Durrett.
\newblock \emph{Probability: Theory and Examples}.
\newblock Duxbury Press, 2004.

\bibitem[{El Ghaoui} and Lebret(1997)]{Ghaoui97}
L.~{El Ghaoui} and H.~Lebret.
\newblock Robust solutions to least-squares problems with uncertain data.
\newblock \emph{SIAM Journal on Matrix Analysis and Applications}, 18:\penalty0
  1035--1064, 1997.

\bibitem[Evgeniou et~al.(2000)Evgeniou, Pontil, and Poggio]{Evgeniou00}
T.~Evgeniou, M.~Pontil, and T.~Poggio.
\newblock Regularization networks and support vector machines.
\newblock In A.~Smola, P.~Bartlett, B.~Sch{\"o}lkopf, and D.~Schuurmans,
  editors, \emph{Advances in Large Margin Classifiers}, pages 171--203,
  Cambridge, MA, 2000. MIT Press.

\bibitem[Globerson and Roweis(2006)]{Globerson06}
A.~Globerson and S.~Roweis.
\newblock Nightmare at test time: Robust learning by feature deletion.
\newblock In \emph{ICML '06: Proceedings of the 23rd International Conference
  on Machine Learning}, pages 353--360, New York, NY, USA, 2006. ACM Press.

\bibitem[Hampel(1974)]{Hample74}
F.~Hampel.
\newblock The influence curve and its role in robust estimation.
\newblock \emph{Journal of the American Statistical Association}, 69\penalty0
  (346):\penalty0 383--393, 1974.

\bibitem[Hampel et~al.(1986)Hampel, Ronchetti, Rousseeuw, and
  Stahel]{HampelRonchettiRousseeumStahel86}
F.~R. Hampel, E.~M. Ronchetti, P.~J. Rousseeuw, and W.~A. Stahel.
\newblock \emph{Robust Statistics: The Approach Based on Influence Functions}.
\newblock John Wiley \& Sons, New York, 1986.

\bibitem[Huber(1981)]{Huber81}
P.~Huber.
\newblock \emph{Robust Statistics}.
\newblock John Wiley \& Sons, New York, 1981.

\bibitem[Kearns et~al.(1997)Kearns, Mansour, Ng, and Ron]{Kearns97}
M.~Kearns, Y.~Mansour, A.~Ng, and D.~Ron.
\newblock An experimental and theoretical comparison of model selection
  methods.
\newblock \emph{Machine Learning}, 27:\penalty0 7--50, 1997.

\bibitem[Koltchinskii and Panchenko(2002)]{Koltchinskii02}
V.~Koltchinskii and D.~Panchenko.
\newblock Empirical margin distributions and bounding the generalization error
  of combined classifiers.
\newblock \emph{The Annals of Statistics}, 30\penalty0 (1):\penalty0 1--50,
  2002.

\bibitem[Kutin and Niyogi(2002)]{KutinNiyogi02}
Samuel Kutin and Partha Niyogi.
\newblock Almost-everywhere algorithmic stability and generalization error.
\newblock In \emph{In UAI-2002: Uncertainty in Artificial Intelligence}, number
  275--282, 2002.

\bibitem[Lanckriet et~al.(2002)Lanckriet, {El Ghaoui}, Bhattacharyya, and
  Jordan]{Lanckriet02}
G.~Lanckriet, L.~{El Ghaoui}, C.~Bhattacharyya, and M.~Jordan.
\newblock A robust minimax approach to classification.
\newblock \emph{Journal of Machine Learning Research}, 3:\penalty0 555--582,
  December 2002.

\bibitem[Maronna et~al.(2006)Maronna, Martin, and Yohai]{MaronnaMartinYohai06}
R.~A. Maronna, R.~D. Martin, and V.~J. Yohai.
\newblock \emph{Robust Statistics. Theory and Methods.}
\newblock John Wiley \& Sons, New York, 2006.

\bibitem[Mukherjee et~al.(2006)Mukherjee, Niyogi, Poggio, and
  Rifkin]{MukherjeeNiyogiPoggioRifkin06}
S.~Mukherjee, P.~Niyogi, T.~Poggio, and R.~Rifkin.
\newblock Learning theory: Stability is sufficient for generalization and
  necessary and sufficient for consistency of empirical risk minimization.
\newblock \emph{Advances in Computational Mathematics}, 25\penalty0
  (1-3):\penalty0 161--193, 2006.

\bibitem[Poggio et~al.(2004)Poggio, Rifkin, Mukherjee, and
  Niyogi]{PoggioRifkinMukherjeeNiyogi04}
T.~Poggio, R.~Rifkin, S.~Mukherjee, and P.~Niyogi.
\newblock General conditions for predictivity in learning theory.
\newblock \emph{Nature}, 428\penalty0 (6981):\penalty0 419--422, 2004.

\bibitem[Rousseeuw and Leeroy(1987)]{RousseeuwLeroy87}
P.~Rousseeuw and A.~Leeroy.
\newblock \emph{Robust Regression and Outlier Detection}.
\newblock John Wiley \& Sons, New York, 1987.

\bibitem[Sch{\"o}lkopf and Smola(2002)]{Scholkopf02}
B.~Sch{\"o}lkopf and A.~Smola.
\newblock \emph{Learning with Kernels}.
\newblock MIT Press, 2002.

\bibitem[Shivaswamy et~al.(2006)Shivaswamy, Bhattacharyya, and
  Smola]{Shivaswamy06}
P.~Shivaswamy, C.~Bhattacharyya, and A.~Smola.
\newblock Second order cone programming approaches for handling missing and
  uncertain data.
\newblock \emph{Journal of Machine Learning Research}, 7:\penalty0 1283--1314,
  July 2006.

\bibitem[Smola et~al.(1998)Smola, Sch{\"o}lkopf, and M{\"u}llar]{Smola98}
A.~Smola, B.~Sch{\"o}lkopf, and K.~M{\"u}llar.
\newblock The connection between regularization operators and support vector
  kernels.
\newblock \emph{Neural Networks}, 11:\penalty0 637--649, 1998.

\bibitem[Steinwart(2005)]{Steinwart05}
I.~Steinwart.
\newblock Consistency of support vector machines and other regularized kernel
  classifiers.
\newblock \emph{IEEE Transactions on Information Theory}, 51\penalty0
  (1):\penalty0 128--142, 2005.

\bibitem[Teo et~al.(2008)Teo, Globerson, Roweis, and
  Smola]{TeoGlobersonRoweisSmola07}
C.~H. Teo, A.~Globerson, S.~Roweis, and A.~Smola.
\newblock Convex learning with invariances.
\newblock In J.C. Platt, D.~Koller, Y.~Singer, and S.~Roweis, editors,
  \emph{Advances in Neural Information Processing Systems 20}, pages
  1489--1496, Cambridge, MA, 2008. MIT Press.

\bibitem[Trafalis and Gilbert(2007)]{Trafalis07}
T.~Trafalis and R.~Gilbert.
\newblock Robust support vector machines for classification and computational
  issues.
\newblock \emph{Optimization Methods and Software}, 22\penalty0 (1):\penalty0
  187--198, February 2007.

\bibitem[{van der Vaart} and Wellner(2000)]{Vaart2000}
A.~{van der Vaart} and J.~Wellner.
\newblock \emph{Weak Convergence and Empirical Processes}.
\newblock Springer-Verlag, New York, 2000.

\bibitem[Vapnik and Chervonenkis(1974)]{Vapnik74}
V.~Vapnik and A.~Chervonenkis.
\newblock \emph{Theory of Pattern Recognition}.
\newblock Nauka, Moscow, 1974.

\bibitem[Vapnik and Chervonenkis(1991)]{Vapnik91}
V.~Vapnik and A.~Chervonenkis.
\newblock The necessary and sufficient conditions for consistency in the
  empirical risk minimization method.
\newblock \emph{Pattern Recognition and Image Analysis}, 1\penalty0
  (3):\penalty0 260--284, 1991.

\bibitem[Vapnik and Lerner(1963)]{Vapnik63}
V.~Vapnik and A.~Lerner.
\newblock Pattern recognition using generalized portrait method.
\newblock \emph{Automation and Remote Control}, 24:\penalty0 744--780, 1963.

\end{thebibliography}

\end{document}